%% file: Zhou-Moosavi-all.tex
\def\year{2018}\relax
\documentclass[letterpaper]{article} 
\usepackage{aaai18}  
\usepackage{times}  
\usepackage{helvet}  
\usepackage{courier}  
\usepackage{url}  
\usepackage{graphicx}  
\usepackage[utf8]{inputenc}
\usepackage{titling}
\usepackage{amsmath,amsthm,amssymb}
\usepackage{thmtools,thm-restate}
\usepackage{outlines}
\usepackage{mathrsfs}
\usepackage{subfigure}
\usepackage{multirow}
\usepackage{algorithm}
\usepackage[noend]{algpseudocode}
\graphicspath{{./figure/}}

\makeatletter
\def\BState{\State\hskip-\ALG@thistlm}
\makeatother

\newcommand{\figurewidthone}{1}
\newcommand{\figurewidthtwo}{0.58}

\DeclareMathOperator*{\argmax}{argmax}
\title{\textbf{Adaptive Quantization for Deep Neural Network}}
\author{\textbf{\Large
Yiren Zhou\textsuperscript{1}, Seyed-Mohsen Moosavi-Dezfooli\textsuperscript{2}, Ngai-Man Cheung\textsuperscript{1}, Pascal Frossard\textsuperscript{2}}\\
\textsuperscript{1}Singapore University of Technology and Design (SUTD)\\
\textsuperscript{2}École Polytechnique Fédérale de Lausanne (EPFL)\\
yiren\_zhou@mymail.sutd.edu.sg, ngaiman\_cheung@sutd.edu.sg\\
\{seyed.moosavi, pascal.frossard\}@epfl.ch\\
}
\date{}
\begin{document}
\maketitle
\begin{abstract}

In recent years Deep Neural Networks (DNNs) have been rapidly developed in various applications, together with increasingly complex architectures. The performance gain of these DNNs generally comes with high computational costs and large memory consumption, which may not be affordable for mobile platforms. Deep model quantization can be used for reducing the computation and memory costs of DNNs, and deploying complex DNNs on mobile equipment. In this work, we propose an optimization framework for deep model quantization. First, we propose a measurement to estimate the effect of parameter quantization errors in individual layers on the overall model prediction accuracy. Then, we propose an optimization process based on this measurement for finding optimal quantization bit-width for each layer. This is the first work that theoretically analyse the relationship between parameter quantization errors of individual layers and model accuracy. Our new quantization algorithm outperforms previous quantization optimization methods, and achieves 20-40\% higher compression rate compared to equal bit-width quantization at the same model prediction accuracy.

\end{abstract}
\section{Introduction} 
\label{sec:introduction}

Deep neural networks (DNNs) have achieved significant success in various machine learning applications, including image classification~\cite{krizhevsky2012imagenet,Simonyan14c,szegedy2015going}, image retrieval~\cite{hoang2017selective,do2016learning}, and natural language processing~\cite{deng2013new}. These achievements come with increasing computational and memory cost, as the neural networks are becoming deeper~\cite{he2016deep}, and contain more filters per single layer~\cite{zeiler2014visualizing}.

While the DNNs are powerful for various tasks, the increasing computational and memory costs make it difficult to apply on mobile platforms, considering the limited storage space, computation power, energy supply of mobile devices~\cite{han2015deep}, and the real-time processing requirements of mobile applications. There is clearly a need to reduce the computational resource requirements of DNN models so that they can be deployed on mobile devices~\cite{7543518}.

In order to reduce the resource requirement of DNN models, one approach relies on model pruning. By pruning some parameters in the model~\cite{han2015deep}, or skipping some operations during the evaluation~\cite{figurnov2016perforatedcnns}, the storage space and/or the computational cost of DNN models can be reduced. Another approach consists in parameter quantization~\cite{han2015deep}. By applying quantization on model parameters, these parameters can be stored and computed under lower bit-width. The model size can be reduced, and the computation becomes more efficient under hardware support~\cite{han2016eie}. It is worth noting that model pruning and parameter quantization can be applied at the same time, without interfering with each other~\cite{han2015deep}; we can apply both approaches to achieve higher compression rates.

Many deep model compression works have also considered using parameter quantization~\cite{gupta2015deep,han2015deep,wu2016quantized} together with other compression techniques, and achieve good results. However, these works usually assign the same bit-width for quantization in the different layers of the deep network. In DNN models, the layers have different structures, which lead to the different properties related to quantization. By applying the same quantization bit-width for all layers, the results could be sub-optimal.
It is however possible to assign different bit-width for different layers to achieve optimal quantization result~\cite{hwang2014fixed}.

In this work, we propose an accurate and efficient method to find the optimal bit-width for coefficient quantization on each DNN layer. Inspired by the analysis in~\cite{NIPS2016_6331}, we propose a method to measure the effect of parameter quantization errors in individual layers on the overall model prediction accuracy. Then, by combining the effect caused by all layers, the optimal bit-width is decided for each layer. By this method we avoid the exhaustive search for optimal bit-width on each layer, and make the quantization process more efficient.
We apply this method to quantize different models that have been pre-trained on ImageNet dataset and achieve good quantization results on all models. Our method constantly outperforms recent state-of-the-art, i.e., the SQNR-based method~\cite{lin2016fixed} on different models, and achieves 20-40\% higher compression rate compared to equal bit-width quantization.
Furthermore, we give a theoretical analysis on how the quantization on layers affects DNN accuracy. To the best of our knowledge, this is the first work that theoretically analyses the relationship between coefficient quantization effect of individual layers and DNN accuracy.

%
\section{Related works} 
\label{sec:related_works}

Parameter quantization has been widely used for DNN model compression~\cite{gupta2015deep,han2015deep,wu2016quantized}. The work in~\cite{gupta2015deep} limits the bit-width of DNN models for both training and testing, and stochastic rounding scheme is proposed for quantization to improve the model training performance under low bit-width. The authors in~\cite{han2015deep} use k-means to train the quantization centroids, and use these centroids to quantize the parameters. The authors in~\cite{wu2016quantized} separate the parameter vectors into sub-vectors, and find sub-codebook of each sub-vectors for quantization. In these works, all (or a majority of) layers are quantized with the same bit-width. However, as the layers in DNN have various structures, these layers may have different properties with respect to quantization.
It is possible to achieve better compression result by optimizing quantization bit-width for each layer.

Previous works have been done for optimizing quantization bit-width for DNN models~\cite{hwang2014fixed,anwar2015fixed,lin2016fixed,sun2016intra}. The authors in~\cite{hwang2014fixed} propose an exhaustive search approach to find optimal bit-width for a fully-connected network. In~\cite{sun2016intra}, the authors first use exhaustive search to find optimal bit-width for uniform or non-uniform quantization; then two schemes are proposed to reduce the memory consumption
during model testing. The exhaustive search approach only works for a relatively small network with few layers, while it is not practical for deep networks. As the number of layers increases, the complexity of exhaustive search increases exponentially. The authors in~\cite{anwar2015fixed} use mean square quantization error (MSQE) ($L_2$ error) on layer weights to measure the sensitivity of DNN layers to quantization, and manually set the quantization bit-width for each layer. The work in~\cite{lin2016fixed} use the signal-to-quantization-noise ratio (SQNR) on layer weights to measure the effect of quantization error in each layer. These MSQE and SQNR are good metrics for measuring the quantization loss on model weights. However, there is no theoretical analysis to show how these measurements relate to the accuracy of the DNN model, but only empirical results are shown. The MSQE-based approach in~\cite{anwar2015fixed} minimizes the $L_2$ error on quantized weight, indicating that the $L_2$ error in different layer has the equal effect on the model accuracy. Similarly, in~\cite{lin2016fixed}, the authors maximize the overall SQNR, and suggest that quantization on different layers has equal contribution to the overall SQNR, thus has equal effect on model accuracy. Both works ignore that the various structure and position of different layers may lead to different robustness on quantization, and thus render the two approaches suboptimal.

In this work, we follow the analysis in~\cite{NIPS2016_6331}, and propose a method to measure the effect of quantization error in each DNN layers. Different from~\cite{anwar2015fixed,lin2016fixed}, which use empirical results to show the relationship between the measurement and DNN accuracy, we conduct a theoretical analysis to show how our proposed method relates to the model accuracy. Furthermore, we show that our bit-width optimization method is more general than the method in~\cite{lin2016fixed}, which makes our optimization more accurate.

There are also works~\cite{hinton2015distilling,romero2014fitnets} that use knowledge distillation to train a smaller network using original complex models. It is also possible to combine our quantization framework with knowledge distillation to achieve yet better compression results.

%
\section{Measuring the effect of quantization noise} 
\label{sec:measuring_the_effect_of_quantization_noise}

In this section, we analyse the effect of quantization on the accuracy of a DNN model. Parameter quantization can result in quantization noise that would affect the performance of the model. Previous works have been done for analyzing the effect of input noise on the DNN model~\cite{NIPS2016_6331}; here we use this idea to analyse the effect of noise in intermediate feature maps in the DNN model.

\subsection{Quantization optimization}
\label{ssec:quantization_optimization}

The goal of our paper is to find a way to achieve optimal quantization result to compress a DNN model. After the quantization, under controlled accuracy penalty, we would like the model size to be as small as possible. Suppose that we have a DNN $\mathcal{F}$ with $N$ layers. Each layer $i$ has $s_i$ parameters, and we apply $b_i$ bit-width quantization in the parameters of layer $i$ to obtain a quantized model $\mathcal{F'}$. Our optimization objective is:

\begin{equation}
\begin{split}
&~min \sum_{i=1}^{N}s_i\cdot b_i \\
s.t.~&~acc_{\mathcal{F}} - acc_{\mathcal{F'}} \leq \Delta_{acc},
\end{split}
\label{eq:optimization_0}
\end{equation}

where $acc_{\mathcal{F}}$ is the accuracy of the model $\mathcal{F}$, and $\Delta_{acc}$ is the maximum accuracy degradation. Note that it requires enormous computation to calculate the accuracy of the model for all quantization cases. To solve the problem more efficiently, we propose a method to estimate the value of the performance penalty given by $acc_{\mathcal{F}} - acc_{\mathcal{F'}}$.

\subsection{Quantization noise}
\label{ssec:quantization_noise}

Value quantization is a simple yet effective way to compress a model~\cite{han2015deep}. Here we evaluate the effect of using value quantization on model parameters.

Assume that conducting quantization on a value is equivalent to adding noise to the value:

\begin{equation}
w_q = w + r_w
\label{eq:quantization_noise}
\end{equation}

Here $w$ is the original value, $w\in W$, with $W$ the set of weights in a layer. Then, $w_q$ is the quantized value, and $r_w$ is the quantization noise.
Assume that we use a uniform quantizer, and that the stepsize of the quantized interval is fixed.
%
Following the uniform quantization analysis in~\cite{you2010audio},
if we consider $\textbf{r}_w=(r_{w,1},\cdots,r_{w,N_W})$ as the quantization noise on all weights in $W$,
we have the expectation of $||\textbf{r}_w||_2^2$ as


\begin{equation}
E(||\textbf{r}_w||_2^2) = p_w'\cdot e^{-\alpha\cdot {b}},
\label{eq:quantization_noise_expectation}
\end{equation}

where $p_w'=N_W\frac{(w_{min}-w_{max})^2}{12}$, $N_W$ is the number of weights in $W$, and $\alpha =ln(4)$~\cite{you2010audio}.
Detailed analysis can be found in Supplementary Material.
Eq.~(\ref{eq:quantization_noise_expectation}) indicates that every time we reduce the bit-width by 1 bit, $E(||\textbf{r}_w||_2^2)$ will increase by 4 times. This is equivalent to the quantization efficiency of 6dB/bit in~\cite{Gray:2006:QUA:2263200.2265587}.



\subsection{Measurement for quantization noise}
\label{ssec:measurement_for_quantization_noise}

\begin{figure}[htbp]
\centering
\includegraphics[width=\figurewidthone\columnwidth]{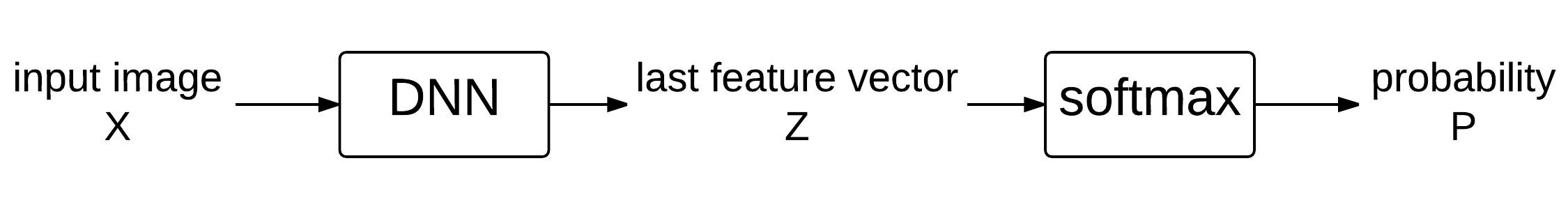}
\caption[]{Simple DNN model architecture.}
\label{fig:model_simple}
\end{figure}

\subsubsection{From weight domain to feature domain}
\label{sssec:from_weight_domain_to_feature_domain}

Eq.~(\ref{eq:quantization_noise_expectation}) shows the quantization noise in the weight domain; here we show how the noise on weight domain can link to the noise in the feature domain.

A simplified DNN classifier architecture is shown in Fig.~\ref{fig:model_simple}.

Here we define $W_i$ as the weights of layer $i$ in the DNN model $\mathcal{F}$. And $Z$ is the last feature map (vector) of the DNN model $\mathcal{F}$. As we quantize $W_i$, the quantization noise is $\textbf{r}_{W_i}$, and there would be a resulting noise $\textbf{r}_{Z_i}$ on the last feature map $Z$.
Here we define $\textbf{r}_{Z_i}$ as the noise on last feature map $Z$ that is caused by the quantization only on a single layer $i$.

As the value of $||\textbf{r}_{Z_i}||_2^2$ is proportional to the value of $||\textbf{r}_{W_i}||_2^2$,
similar to Eq.~(\ref{eq:quantization_noise_expectation}), the expectation of resulting noise on $\textbf{r}_{Z_i}$ is:

\begin{equation}
E(||\textbf{r}_{Z_i}||_2^2) = p_i\cdot e^{-\alpha\cdot {b_i}}
\label{eq:quantization_noise_feature_expectation}
\end{equation}

This is proved in later sections, empirical results are shown in Fig.~\ref{fig:weight_fmlast}.

\subsubsection{The effect of quantization noise}
\label{sssec:the_effect_of_quantization_noise}

Similarly to the analysis in~\cite{pang2017robust}, we can see that the softmax classifier has a
linear decision boundary in the last feature vectors $Z$~\cite{pang2017robust} in Fig.~\ref{fig:model_simple}.
The analysis can be found in the Supplementary Material.
Then we apply the result of~\cite{NIPS2016_6331} to bound the robustness of the classifier with respect to manipulation of weights in different layers.

We define $\textbf{r}^*$ to be the adversarial noise, which represents the minimum noise to cause misclassification. For a certain input vector $\textbf{z}=(z_1,\cdots,z_L)$, where $L$ is the number of element in $\textbf{z}$, the $||\textbf{r}^*||_2$ is the distance from the datapoint to the decision boundary, which is a fixed value. We define a sorted vector of $\textbf{z}$ as $\textbf{z}_{sorted}=(z_{(1)},\cdots,z_{(L)})$, where the max value is $z_{(1)}$, and second max value is $z_{(2)}$. The result for softmax classifier (or max classifier) can be expressed as: $max(\textbf{z})=z_{(1)}$, which is picking up the maximum value in the vector$\textbf{z}$.

As adversarial noise is the minimum noise that can change the result of a classifier, we can get the adversarial noise for softmax classifier $max(\textbf{z})$ as $\textbf{r}_{\textbf{z}_{sorted}}^*=(\frac{z_{(2)}-z_{(1)}}{2},\frac{z_{(1)}-z_{(2)}}{2},0,\cdots,0)$, then the norm square of adversarial noise $||\textbf{r}^*||_2^2=(z_{(1)}-z_{(2)})^2/2$.

Here we define $\textbf{r}_Z$ as the noise that we directly add on last feature map $Z$. We can consider $\textbf{r}_Z$ as the collective effect of all $\textbf{r}_{Z_i}$ that caused by the quantization on all layers $i\in\{1,\cdots,N\}$, where $N$ is the number of layers.

As mentioned in~\cite{NIPS2016_6331}, if we apply random noise $\|\textbf{r}_Z\|_2^2$ rather than adversarial noise $\|\textbf{r}^*\|_2$ on the input vector $\textbf{z}$ for a softmax classifier $max(\textbf{z})$, it requires higher norm for random noise $\|\textbf{r}_Z\|_2^2$ to causes prediction error with same probability, compared to adversarial noise $\|\textbf{r}^*\|_2$.

The following result shows the relationship between the random noise $\|\textbf{r}_Z\|_2^2$ and adversarial noise $\|\textbf{r}^*\|_2$, under softmax classifier with a number of classes equal to $d$:

\begin{restatable}{lemma}{primelemma}
Let $\gamma(\delta)=5+4\ln(1/\delta)$. The following inequalities hold between the norm square of random noise $\|\textbf{r}_Z\|_2^2$ and adversarial noise $\frac{(z_{(1)}-z_{(2)})^2}{2}$.
\begin{equation}
\frac{\ln{d}}{d}\gamma(\delta) \|\textbf{r}_Z\|_2^2 \geq
\frac{(z_{(1)}-z_{(2)})^2}{2}
\end{equation}
with probability exceeding $1 - 2 \delta$.
\label{lem:linear}
\end{restatable}

The proof of Lemma~\ref{lem:linear} can be found in the Supplementary Material.
The lemma states that if the norm of random noise is $o\left((z_{(1)}-z_{(2)})\sqrt{d/\ln d}\right)$, it does not change the classifier decision with high probability.

Based on Lemma~\ref{lem:linear}, we can rewrite our optimization problem. Assume that we have a model $\mathcal{F}$ with accuracy $acc_{\mathcal{F}}$. After adding random noise $\textbf{r}_Z$ on the last feature map $Z$, the model accuracy drops by $\Delta_{acc}$. If we have

\begin{equation}
\theta(\Delta_{acc}) = \frac{d}{\gamma(\frac{\Delta_{acc}}{2acc_{\mathcal{F}}})\ln{d}},
\label{eq:theta_define}
\end{equation}

we have the relation between accuracy degradation and noise $\textbf{r}_Z$ as:

\begin{equation}
\begin{split}
&~acc_{\mathcal{F}} - acc_{\mathcal{F'}} \leq \Delta_{acc} \Rightarrow \\
&~\|\textbf{r}_Z\|_2^2 < \theta(\Delta_{acc})\frac{(z_{(1)}-z_{(2)})^2}{2}
\end{split}
\label{eq:relation_acc_noise}
\end{equation}

The detailed analysis can be found in the Supplementary Material.
Eq.~(\ref{eq:relation_acc_noise}) shows the bound of noise on last feature map $Z$. However, adding quantization noise to different layers may have different effect on model accuracy.
Suppose we have model $\mathcal{F}$ for quantization. By adding noise $\textbf{r}_{W_i}$ on weights of layer $i$, we induce the noise $\textbf{r}_{Z_i}$ on last feature map $Z$. By quantizing earlier layers, the noise needs to pass through more layers to get $\textbf{r}_{Z_i}$, which results in a low rank noise $\textbf{r}_{Z_i}$. For example, when quantizing the first layer, is results in $\textbf{r}_{Z_1}=(e_1,\cdots,e_{d'},0,\cdots,0)$, and $rank(\textbf{r}_{Z_1})=d'<d$. When quantizing the last layer, it results in $\textbf{r}_{Z_N}=(e_1,\cdots,e_{d})$, and $rank(\textbf{r}_{Z_N})=d$. In order to let $\textbf{r}_{Z_N}$ have equivalent effect on model accuracy as $\textbf{r}_{Z_1}$, $\|\textbf{r}_{Z_N}\|_2$ should be larger than $\|\textbf{r}_{Z_1}\|_2$.

By considering the different effects of $\textbf{r}_{Z_i}$ caused by quantization in different layers, we rewrite Eq.~(\ref{eq:relation_acc_noise}) in a more precise form:

\begin{equation}
\begin{split}
&~acc_{\mathcal{F}} - acc_{\mathcal{F'}} = \Delta_{acc} \Rightarrow \\
&~\|\textbf{r}_{Z_i}\|_2^2 = t_i(\Delta_{acc})\frac{(z_{(1)}-z_{(2)})^2}{2}.
\end{split}
\label{eq:relation_acc_noise_precise}
\end{equation}

Here $t_i(\Delta_{acc})$ is the robustness parameter of layer $i$ under accuracy degradation $\Delta_{acc}$.

Eq.~(\ref{eq:relation_acc_noise_precise}) shows a precise relationship between $\textbf{r}_{Z_i}$ and $\Delta_{acc}$. If we add quantization noise to layer $i$ of model $\mathcal{F}$, and get noise $\textbf{r}_{Z_i}$ on last feature map $Z$, then the model accuracy decreases by $\Delta_{acc}$.


We consider the layer $i$ in model $\mathcal{F}$ as $\textbf{y}_i=\mathcal{F}_i(\textbf{y}_{i-1})$, where $\textbf{y}_i$ is the feature map after layer $i$. Here we consider that the noise $\textbf{r}_{y_i}$ would transfer through layers under almost linear transformation (to be discussed in later sections). If we add random noise in the weights of layer $i$, we have the rank of the resulting noise $\textbf{r}_{Z_i}$ on last feature map $Z$ given as:

\begin{equation}
rank(\textbf{r}_{Z_i}) = rank(\prod_{i}^{N}\mathcal{F}_i) \leq min\{rank(\mathcal{F}_i)\}
\label{eq:rank_noise_multi}
\end{equation}

Based on Eq.~(\ref{eq:rank_noise_multi}), we have:

\begin{equation}
rank(\textbf{r}_{Z_1}) \leq \cdots \leq rank(\textbf{r}_{Z_i}) \leq \cdots \leq rank(\textbf{r}_{Z_N})
\label{eq:rank_noise_seq}
\end{equation}

Eq.~(\ref{eq:rank_noise_seq}) suggests that the noise on earlier layers of DNN needs to pass through more layers to affect the last feature map $Z$, the noise $\textbf{r}_{Z_i}$ on $Z$ would have lower rank, resulting in a lower value of $t_i(\Delta_{acc})$.


From Eq.~(\ref{eq:relation_acc_noise_precise}), we can see in particular that when

\begin{equation}
\frac{\|\textbf{r}_{Z_i}\|_2^2}{t_i(\Delta_{acc})} = \frac{\|\textbf{r}_{Z_j}\|_2^2}{t_j(\Delta_{acc})},
\label{eq:robustness_equal}
\end{equation}

the quantization on layer $i$ and $j$ have same effect on model accuracy. Based on Eq.~(\ref{eq:robustness_equal}), $\frac{\|\textbf{r}_{Z_i}\|_2^2}{t_i(\Delta_{acc})}$ can be a good measurement for estimating the accuracy degradation caused by quantization noise, regardless of which layer to quantize.
Consider $\textbf{x}\in \mathcal{D}$ as the input in dataset $\mathcal{D}$, we have the corresponding feature vector $\textbf{z}=\mathcal{G}(\textbf{x},W)$ in the last feature map $Z$. By quantizing layer $i$ in model $\mathcal{F}$, we get noise $\textbf{r}_{\textbf{z}_i}$ on $\textbf{z}$. We define the accuracy measurement on layer $i$ as:

\begin{equation}
m_i=\frac{\frac{1}{|\mathcal{D}|}\sum_{\textbf{x}\in \mathcal{D}}(||\textbf{r}_{\textbf{z}_i}||_2^2)}{t_i(\Delta_{acc})}=\frac{||\textbf{r}_{Z_i}||_2^2}{t_i(\Delta_{acc})}
\label{eq:measurement_layer}
\end{equation}

The way to calculate $t_i(\Delta_{acc})$ is given by:

\begin{equation}
\begin{split}
t_i(\Delta_{acc}) = \frac{mean_{\textbf{r}_{\textbf{z}_i}}}{mean_{\textbf{r}^*}}
= \frac{\frac{1}{|\mathcal{D}|}\sum_{\textbf{x}\in \mathcal{D}}||\textbf{r}_{\textbf{z}_i}||_2^2}{\frac{1}{|\mathcal{D}|}\sum_{\textbf{x}\in \mathcal{D}}\frac{(z_{(1)}-z_{(2)})^2}{2}} \\
s.t.~~~acc_{\mathcal{F}} - acc_{\mathcal{F'}} = \Delta_{acc} 
\end{split}
\label{eq:calculate_t}
\end{equation}

The detailed method to calculate $t_i(\Delta_{acc})$ will be discussed in the experiment section. Note that, based on the optimization result in Eq.~(\ref{eq:optimization_result}), the selected value of $\Delta_{acc}$ does not matter for the optimization result, as long as the the value of $\frac{t_i(\Delta_{acc})}{t_j(\Delta_{acc})}$ is almost independent w.r.t. $\Delta_{acc}$, which is true according to Fig.~\ref{fig:fm_last_top1_noise}. So choosing different value of $\Delta_{acc}$ does not change the optimization result. In later sections, we use $t_i$ instead of $t_i(\Delta_{acc})$ for simplicity.

From Eq.~(\ref{eq:measurement_layer}), based on the linearity and additivity of the proposed estimation method (shown in later sections), the measurement of the effect of quantization error in all the layers of the DNN model is shown in Eq.~(\ref{eq:measurement_all}).

After we define the accuracy measurement for each layer of model, based on Eq.~(\ref{eq:relation_acc_noise_precise}), we can then rewrite the optimization in Eq.~(\ref{eq:optimization_0}) as

\begin{equation}
\begin{split}
&~min \sum_{i=1}^{N}s_i\cdot b_i \\
s.t.~&~m_{all} = \sum_{i=1}^{N}m_i \leq C,
\end{split}
\label{eq:optimization_0_change_1}
\end{equation}

where $m_{all}$ is the accuracy measurement for all layers, and $C$ is a constant related to model accuracy degradation $\Delta_{acc}$, with higher $C$ indicating higher $\Delta_{acc}$.

\subsection{Linearity of the measurements}
\label{ssec:linearity_of_the_measurements}

In this section we will show that the DNN model are locally linear to the quantization noise measurement $||\textbf{r}_W||_2$, under the assumption that the quantization noise is much smaller than the original value: $||r_w||_2 \ll ||w||_2$. That is, if a quantization noise $||\textbf{r}_{W_i}||_2$ on layer $i$ leads to $||\textbf{r}_{Z_i}||_2$ on last feature vector $Z$, then we have a quantization noise $\alpha\cdot ||\textbf{r}_{W_i}||_2$ on layer $i$ leads to $\alpha\cdot||\textbf{r}_{Z_i}||_2$ on last feature vector $Z$.

For linear layers like convolutional layers and fully connected layers in the DNN model, the linearity for noise is obvious. Here we mainly focus on the non-linear layers in the DNN model, such as ReLU
and Max-pooling layers.

\subsubsection{ReLU layers}

The ReLU layers is widely used to provide nonlinear activation for DNN. Given the input $a\in A$ to a ReLU layer, the output value $z\in Z$ is calculated as:

\begin{equation}
z = ReLU(a) = \left\{\begin{matrix}
a, & if~a>0\\ 
0, & if~a<=0
\end{matrix}\right.
\label{eq:relu}
\end{equation}

From Eq.~(\ref{eq:relu}) we can see that the ReLU layer is linear to noise in most cases. The non-linearity happens only when the noise $r_w$ crossing the zero point, which has small probability when the noise is sufficiently small.

\subsubsection{Max-pooling layers}

Max-pooling is a nonlinear downsampling layer that reduces the input dimension and controls overfitting. We can consider that max-pooling acts as a $max$ filter to the feature maps.

Similarly to the ReLU layer, which can be described as $z=ReLU(a)=max(0,a)$, the max-pooling layer can be describes as $z=maxpool(\{a_i\})=max(\{a_i\})$, where $i=1,\cdots,P$, with $P$ the kernel size for max-pooling. The linearity for noise holds when the noises are sufficiently small and do not alter the order for $\{a_i\}$.

\subsubsection{Other layers}

For other non-linear layers like Sigmoid and PReLU, the linearity for small noises still holds under the assumptions that the function is smooth along most input ranges, and the noise has very low probability to cross the non-linear region.

Based on the linearity assumption, as we model the quantization noise on weight as Eq.~(\ref{eq:quantization_noise_expectation}), the resulting noise on last feature vector $Z$ can be modeled as:

\begin{equation}
||\textbf{r}_{Z_i}||_2^2=p_i\cdot e^{-\alpha\cdot b_i}
\label{eq:noise_single}
\end{equation}

\subsection{Additivity of the measurements}
\label{ssec:additivity_of_the_measurements}

\subsubsection{Noise on single multiplication}

Pairwise multiplication is a basic operation in the convolutional layers and fully connected layers of DNN. Given one value in the input $a\in A$, one value in the weight matrix $w\in W$, we have the pairwise multiplication as $a\cdot w$.
%
%
If we consider noise in both input $a\in A$ and $w\in W$, we have noised value $a_q\in A_q$ and $w_q\in W_q$,
and finally $a_q\cdot w_q = (a+r_a)\cdot (w+r_w)$.


\subsubsection{Noise on one layer}
\label{sssec:noise_on_one_layer}

Given a convolutional layer input $A$ with size $M\times N\times C$, conv kernel $K$ with size $M_k\times N_k\times C\times D$, and stride size $s$, we have the output feature map $Z$ with size $(M/s)\times (N/s)\times D$. Here $M$ and $N$ are the height and width of input, $C$ is the number of channels of input. $M_k$ and $N_k$ are the height and width of the conv kernel, $D$ is the depth of output feature map.

The analysis on fully connected layers will be similar to the analysis on convolutional layers. It can be considered as a special case of convolutional layers when $M$, $N$, $M_k$, and $N_k$ are equal to 1. For a single value $z_{p, q, d}\in Z$, the noise term of $z_{p, q, d}$ can be expressed as:







\begin{equation}
\begin{split}
r_{z_{p, q, d}} & = \sum_{i=1}^{M_k\cdot N_k\cdot C}(a_i\cdot r_{w_{i, d}} + r_{a_i}\cdot w_{i, d} + r_{a_i}\cdot r_{w_{i, d}}) + r_{b_d} \\
                        & \approx \sum_{i=1}^{M_k\cdot N_k\cdot C}(a_i\cdot r_{w_{i, d}} + r_{a_i}\cdot w_{i, d})
\end{split}
\label{eq:convolution_value_noise_both_2}
\end{equation}

The calculation details can be found in Supplementary Material.
Note that the term $r_{b_d}$ can be ignored under the assumption that $w$ and $b$ have same bit-width quantization. The term $r_{a_i}\cdot r_{w_{i, d}}$ can be ignored under the assumption that $||r_a||_2\ll ||a||_2$ and $||r_w||_2\ll ||w||_2$.

From Eq.~(\ref{eq:convolution_value_noise_both_2}) we can see that: 1) adding noise to input feature maps and weights separately and independently, is equivalent to adding noise to both input feature maps and weights; 2) regarding the output feature map $\textbf{z}\in Z$, adding noise to the input feature maps and weights and doing layer operation (pairwise product), is equivalent to adding the noise directly to the output feature map. We will use these two properties in later sections.

\subsubsection{Adding noise to multiple layers}

\begin{figure}[htbp]
\centering
\subfigure[]{\includegraphics[width=\figurewidthone\columnwidth]{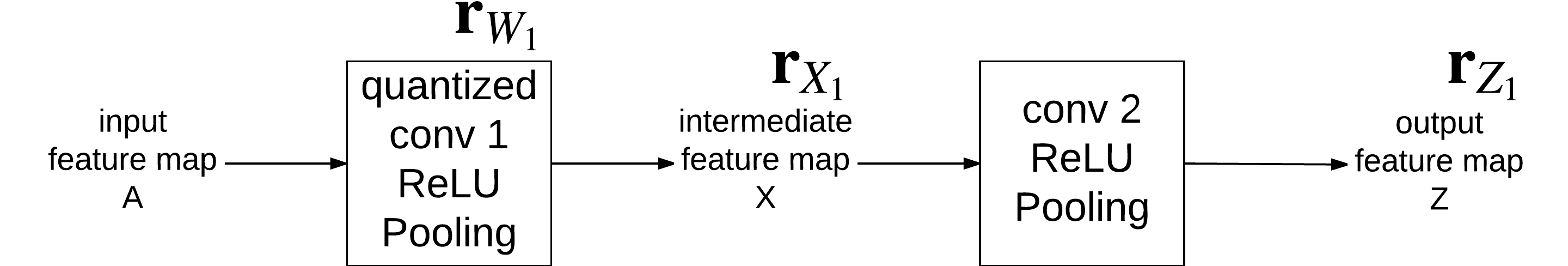}
\label{sfig:model_layer_1}
}
\subfigure[]{\includegraphics[width=\figurewidthone\columnwidth]{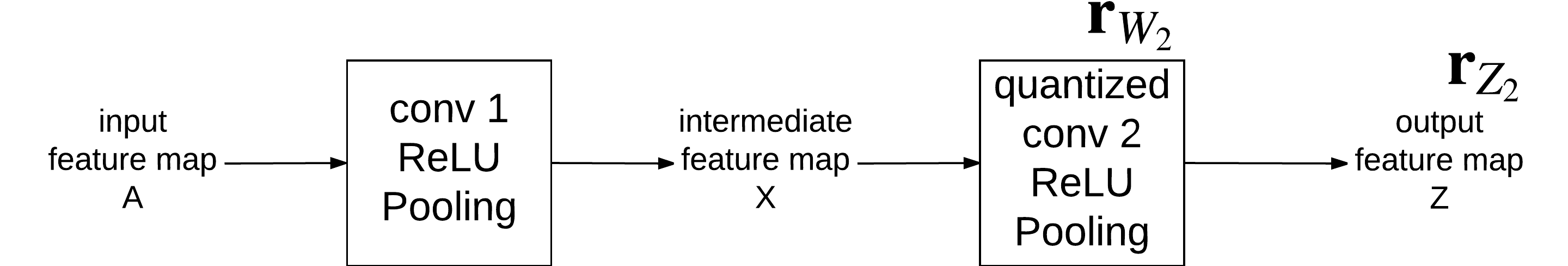}

\label{sfig:model_layer_2}
}
\subfigure[]{\includegraphics[width=\figurewidthone\columnwidth]{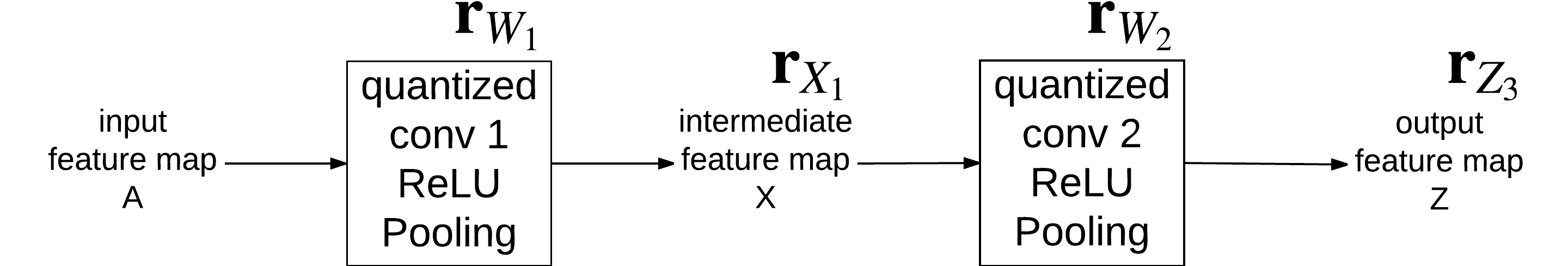}
\label{sfig:model_layer_3}
}
\caption[]{Effect of adding noise to multiple layers.}
\label{fig:model_layer}
\end{figure}

Fig.~\ref{fig:model_layer} shows a 2-layer module inside a DNN model. Given input feature map $A$, after the first conv layer, an intermediate feature map $X$ is generated, then after the second conv layer, output feature map $Z$ is generated.
Fig.~\ref{sfig:model_layer_1} and~\ref{sfig:model_layer_2} show the effect of noise on layer 1 and 2, respectively. And Fig.~\ref{sfig:model_layer_3} shows the effect of noise on both layer 1 and 2. By analysing the additivity of $\|\textbf{r}_{Z}\|_2^2$, we have:

\begin{equation}
||\textbf{r}_{\textbf{z}_3}||_2^2 \doteq ||\textbf{r}_{\textbf{z}_1}||_2^2 + ||\textbf{r}_{\textbf{z}_2}||_2^2
\label{eq:two_layer_add}
\end{equation}

Detailed analysis can be found in Supplementary Material.
Eq.~(\ref{eq:two_layer_add})
holds under the assumption that $\textbf{r}_{\textbf{z}_1}$ and $\textbf{r}_{\textbf{z}_2}$ are independent. This is reasonable in our case, as $\textbf{r}_{\textbf{z}_1}$ and $\textbf{r}_{\textbf{z}_2}$ are caused by $\textbf{r}_{W_1}$ and $\textbf{r}_{W_2}$ which are two independent quantization noises. This independence between $\textbf{r}_{\textbf{z}_1}$ and $\textbf{r}_{\textbf{z}_2}$ is also important for our proposed estimation method.

We can extend Eq.~(\ref{eq:two_layer_add}) to the situation of $N$ layers:

\begin{equation}
||\textbf{r}_{\textbf{z}}||_2^2 = \sum_{i=1}^{N}||\textbf{r}_{\textbf{z}_i}||_2^2
\label{eq:multiple_layer_add}
\end{equation}

If we consider the linearity and additivity of the proposed measurement, from Eq.~(\ref{eq:measurement_layer}) and Eq.~(\ref{eq:multiple_layer_add}), as well as the independence of the measurement among different layers, we have the measurement of the effect of quantization errors in all layers in DNN model:

\begin{equation}
m_{all}=\sum_{i=1}^{N}m_i=\sum_{i=1}^{N}\frac{||\textbf{r}_{Z_i}||_2^2}{t_i}
\label{eq:measurement_all}
\end{equation}

Eq.~(\ref{eq:measurement_all}) suggests that the noise effect of adding noise to each layer separately and independently, is equivalent to the effect of adding noise to all layers simultaneously. We use Eq.~(\ref{eq:measurement_layer}) as the measurement for noise effect on layer $i$, and the effect of adding noise to all layers can be predicted using Eq.~(\ref{eq:measurement_all}).

%
\section{Layer-wise bit-width optimization} 
\label{sec:layer_wise_bit_width_optimization}

In this section we show the approach for optimizing the layer-wise bit-width quantization to achieve an optimal compression ratio under certain accuracy loss.

Following the discussion from the optimization problem in Eq.~(\ref{eq:optimization_0_change_1}), our goal is to constraint Eq.~(\ref{eq:measurement_all}) to be a small value while minimizing the model size.

\subsection{Adaptive quantization on multiple layers}
\label{ssec:adaptive_quantization_on_multiple_layers}



Based on Eq.~(\ref{eq:noise_single}) and~(\ref{eq:measurement_all}),
%
%
the optimization Eq.~(\ref{eq:optimization_0_change_1}) can be expressed as:

\begin{equation}
\begin{split}
&~min \sum_{i=1}^{N}s_i\cdot b_i \\
s.t.& \sum_{i=1}^{N}\frac{p_i}{t_i}\cdot e^{-\alpha \cdot b_i}\leq C
\end{split}
\label{eq:optimization_1}
\end{equation}





The optimal value of Eq.~(\ref{eq:optimization_1}) can be reached when:

\begin{equation}
\frac{p_1\cdot e^{-\alpha \cdot b_1}}{t_1 \cdot s_1}=\frac{p_2\cdot e^{-\alpha \cdot b_2}}{t_2 \cdot s_2}=\cdots=\frac{p_N\cdot e^{-\alpha \cdot b_N}}{t_N \cdot s_N}
\label{eq:optimization_result}
\end{equation}

The detailed analysis can be found in Supplementary Material.

\subsection{Optimal bit-width for each layer}
\label{ssec:optimal_bit-width_for_each_layer}

From Eq.~(\ref{eq:optimization_result}) we can directly find the optimal bit-width for each layer using the following procedure:

\begin{outline}
\1 Calculate $t_i$~(Eq.~(\ref{eq:calculate_t})):
 \2 First, calculate the mean value of adversarial noise for the dataset: $mean_{\textbf{r}^*}=\frac{1}{|\mathcal{D}|}\sum_{\textbf{x}\in \mathcal{D}}\frac{(z_{(1)}-z_{(2)})^2}{2}$.
 \2 Then, fix $\Delta_{acc}$ value. For example, $\Delta_{acc} = 10\%$. Note that the selection of $\Delta_{acc}$ value does not affect the optimization result.
 \2 For each layer $i$, change the amount of noise $\textbf{r}_{W_i}$ added in weight $W_i$, until the accuracy degradation equals to $\Delta_{acc}$. Then, record the mean value of noise $\textbf{r}_{\textbf{z}_i}$ on the last feature map $Z$: $mean_{\textbf{r}_{\textbf{z}_i}}=\frac{1}{|\mathcal{D}|}\sum_{\textbf{x}\in \mathcal{D}}||\textbf{r}_{\textbf{z}_i}||_2^2$.
 \2 The value $t_i$ can be calculated as: $t_i(\Delta_{acc}) = \frac{mean_{\textbf{r}_{\textbf{z}_i}}}{mean_{\textbf{r}^*}}$.
 \2 The details for the calculation of $t_i$ can be found in Fig.~\ref{fig:fm_last_top1_noise}.
\1 Calculate $p_i$:
 \2 First, for each layer $i$, fix $b_i$ value. For example, use $b_i=10$.
 \2 Then, record the mean value of noise $\textbf{r}_{\textbf{z}_i}$ on the last feature map $Z$: $mean_{\textbf{r}_{\textbf{z}_i}}=\frac{1}{|\mathcal{D}|}\sum_{\textbf{x}\in \mathcal{D}}||\textbf{r}_{\textbf{z}_i}||_2^2$.
 \2 The value $p_i$ can be calculated using Eq.~(\ref{eq:noise_single}): $mean_{\textbf{r}_{\textbf{z}_i}}=||\textbf{r}_{Z_i}||_2^2=p_i\cdot e^{-\alpha \cdot b_i}$.
\1 Calculate $b_i$:
 \2 Fix the bitwidth for first layer $b_1$, for example, $b_1=10$. Then bitwidth for layer $i$ can be calculated using the Eq.~(\ref{eq:optimization_result}): $\frac{p_1\cdot e^{-\alpha \cdot b_1}}{t_1 \cdot s_1}=\frac{p_i\cdot e^{-\alpha \cdot b_i}}{t_i \cdot s_i}$
\end{outline}

The detailed algorithm about the above procedure can be found in Supplementary Material. Note that, by selecting different $b_1$, we achieve different quantization result. A lower value of $b_1$ results in higher compression rate, as well as higher accuracy degradation.

\subsection{Comparison with SQNR-based approach}
\label{ssec:comparison_with_SQNR-based_approach}

Based on the SQNR-based approach~\cite{lin2016fixed}, the optimal bit-width is reached when:

\begin{equation}
\frac{e^{-\alpha \cdot b_1}}{s_1}=\frac{e^{-\alpha \cdot b_2}}{s_2}=\cdots=\frac{e^{-\alpha \cdot b_N}}{s_N}
\label{eq:optimization_result_SQNR}
\end{equation}

The proof can be found in Supplementary Material. Note that compared with our result in Eq.~(\ref{eq:optimization_result}), the parameters $p_i$ and $t_i$ are missing. This is consistent with the assumption of the SQNR-based approach, where two layers having the same bit-width for quantization would have the same SQNR value; hence the effects on accuracy are equal. This makes the SQNR-based approach a special case of our approach, when all layers in the DNN model have the equal effect on model accuracy under the same bit-width.

%
\section{Experimental results} 
\label{sec:experimental_results}

In this section we show empirical results that validate our assumptions in previous sections, and evaluate the proposed bit-width optimization approach.

All codes are implemented using MatConvNet~\cite{vedaldi15matconvnet}. All experiments are conducted using a Dell workstation with E5-2630 CPU and Titan X Pascal GPU.

\subsection{Empirical results about measurements}
\label{ssec:empirical_results_about_measurements}

To validate the effectiveness of the proposed accuracy estimation method, we conduct several experiments. These experiments validate the relationship between the estimated accuracy, the linearity of the measurement, and the additivity of the measurement.

Here we use Alexnet~\cite{krizhevsky2012imagenet}, VGG-16~\cite{Simonyan14c}, GoogleNet~\cite{szegedy2015going}, and Resnet~\cite{he2016deep} as the model for quantization. Each layer of the model is quantized separately using uniform quantization, but possibly with different bit-width. The quantized model is then tested on the validation set of Imagenet~\cite{krizhevsky2012imagenet}, which contains 50000 images in 1000 classes.

\subsubsection{Calculate $t_i$}
\label{sssec:calculate_t_i}

As Eq.~(\ref{eq:measurement_layer}) is proposed to measure the robustness of each layer, we conduct an experiment to find $t_i$ value. We use Alexnet as an example.

First, we calculate the adversarial noise for Alexnet on the last feature vector $Z$.
The calculation is based on Eq.~(\ref{eq:calculate_t}). The mean value of $||\textbf{r}^*||_2^2$ for Alexnet is $mean_{\textbf{r}^*}=5.33$. The distribution of $||\textbf{r}^*||_2^2$ for Alexnet on Imagenet validation set can be found in Supplementary Material.

\begin{figure}[!t]
\centering
\subfigure[]{
    \includegraphics[width=\figurewidthtwo\columnwidth]{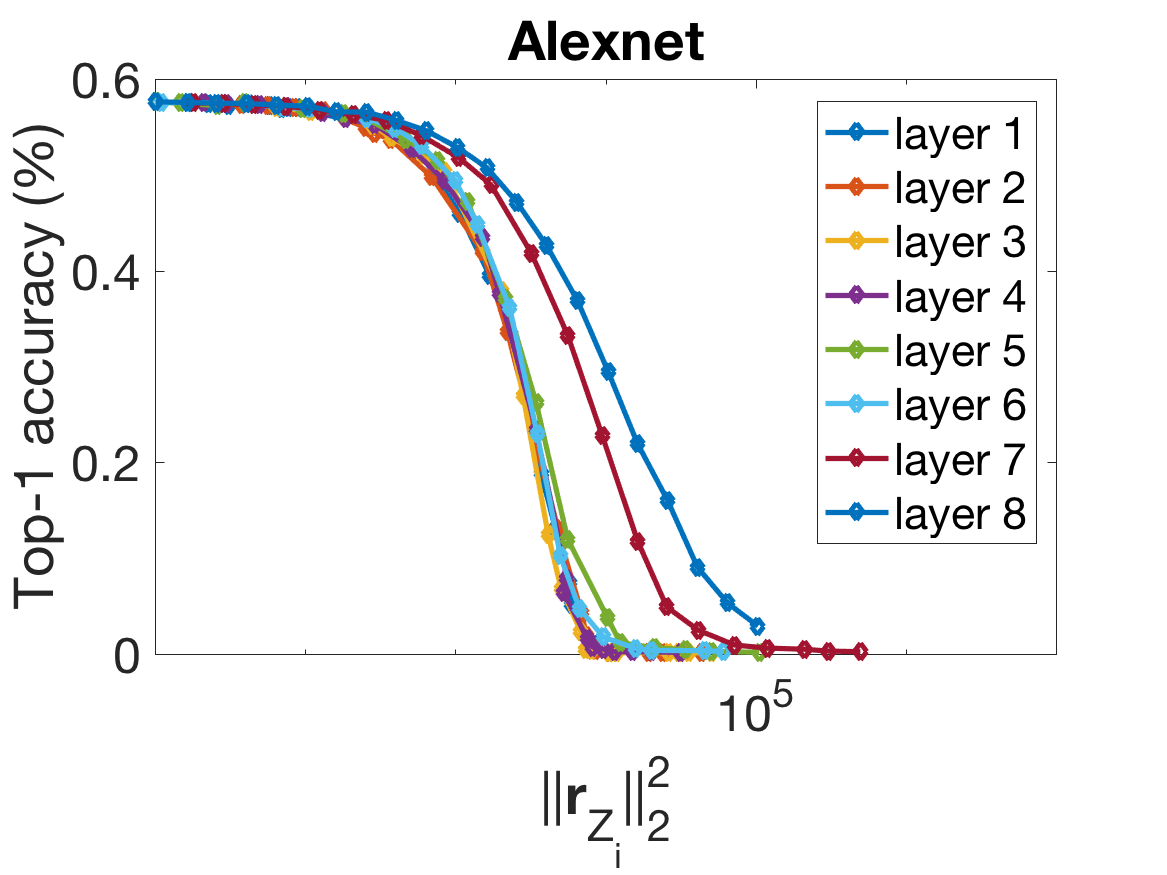}
    \label{fig:fm_last_top1_noise_alex}
    
}
\subfigure[]{
    \includegraphics[width=\figurewidthtwo\columnwidth]{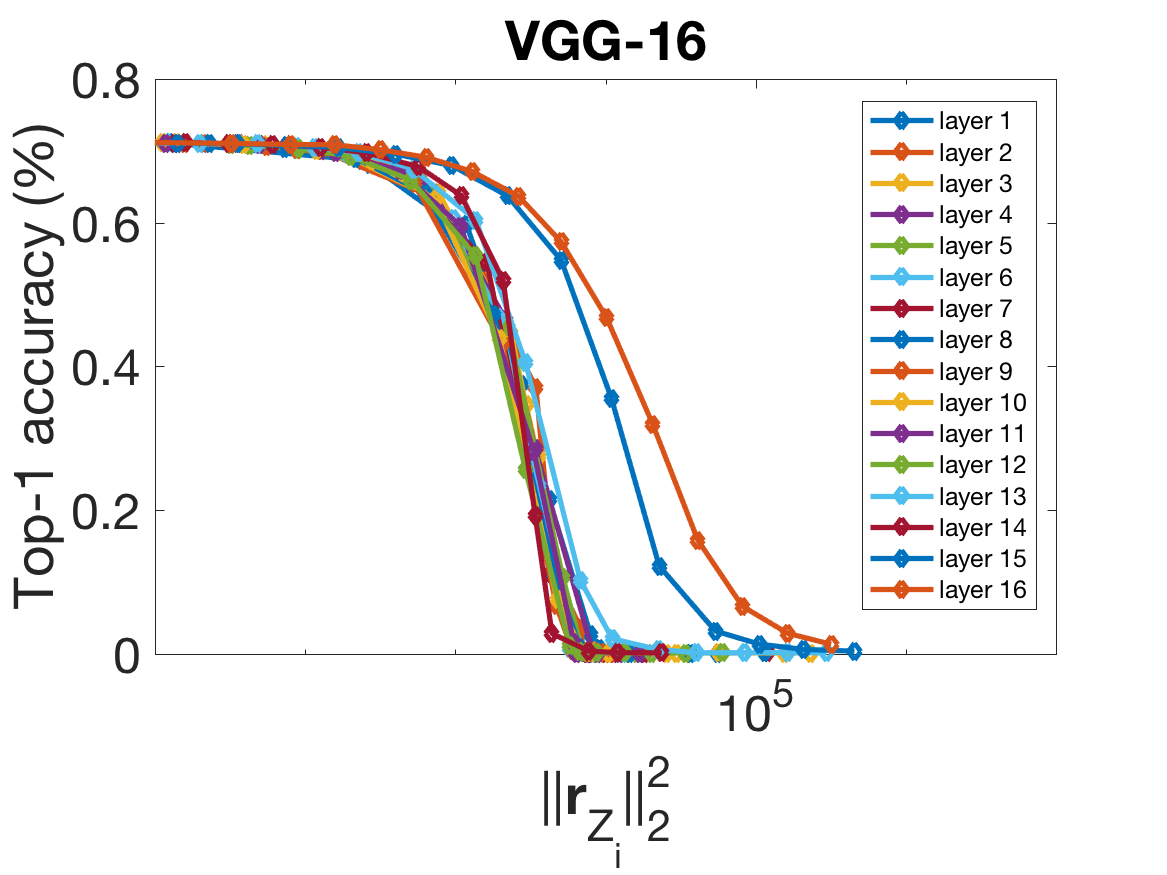}
    \label{fig:fm_last_top1_noise_vgg}
    
}
\caption[]{The relationship between different $||\textbf{r}_{Z_i}||_2^2$ and model accuracy.}
\label{fig:fm_last_top1_noise}

\end{figure}

After finding $||\textbf{r}^*||_2^2$ value, the value of $t_i$ is calculated based on Fig.~\ref{fig:fm_last_top1_noise_alex} and Eq.~(\ref{eq:calculate_t}). We set the accuracy degradation to be roughly half of original accuracy (57\%), which is $28\%$. Based on the values in Fig.~\ref{fig:fm_last_top1_noise_alex}, $t_1$ to $t_6$ are equal to
$5.2\times 10^2$, $t_7=1.3\times 10^3$, and $t_8=2.0\times 10^3$.

Here we show the example for Alexnet of how to calculate the $t_i$ value. Note that for other networks like VGG-16, GoogleNet, and Resnet, we also observe that only the $t_i$ value for the last 1 or 2 layers are obviously different than the other $t_i$ values. During our calculation, we can thus focus on the $t_i$ values for the last several layers. Furthermore, in Fig.~\ref{fig:fm_last_top1_noise_alex}, we find the $||\textbf{r}_Z||_2^2-accuracy$ relationship for different amounts of noise, which requires a lot of calculations. In real cases, we use binary search to find appropriate points under the same accuracy degradation. This makes the process to calculate $t_i$ fast and efficient. Typically, for a deep model with $N$ layers, and dataset with $|\mathcal{D}|$ size, we require $O(\tau N|\mathcal{D}|)$ forward passes to calculate accuracy. Here $\tau$ is the trial times over one layer. We can reduce it to $O(\tau N'|\mathcal{D}|)$(with $N'<<N$) by only calculating $t_i$ values for the last $N'$ layers.

In our experiments, the calculation of $t_i$ is the most time-consuming part of our algorithm.
We use around 15 mins to calculate the $t_i$ value for Alexnet (30 sec for forward pass on the whole dataset), and around 6 hours to calculate the $t_i$ value for Resnet-50 (2 min for forward pass on the whole dataset). This time can be reduced if we only calculate $t_i$ values for the last few layers.


\subsubsection{Linearity of measurements}
\label{sssec:linearity_of_measurements}

\begin{figure}[!t]
\centering
\subfigure[]{
    \includegraphics[width=\figurewidthtwo\columnwidth]{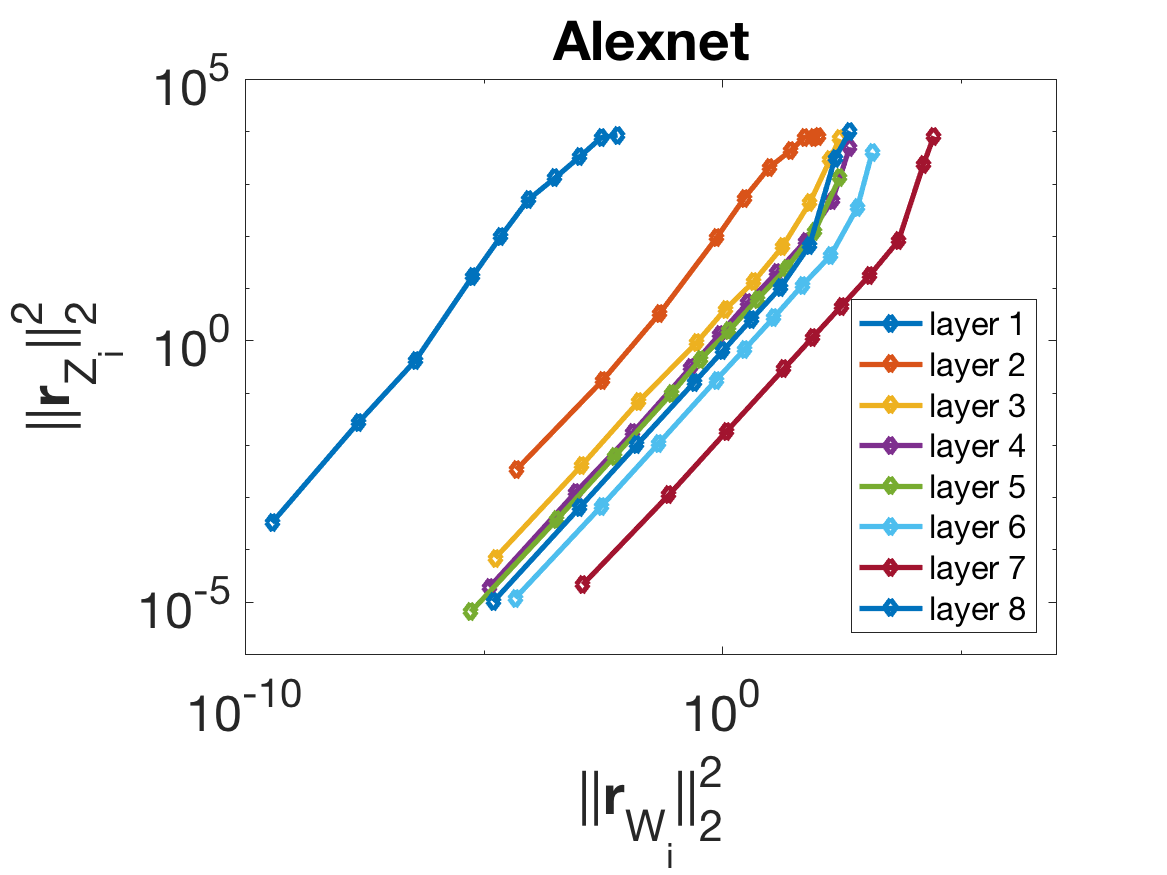}
    \label{fig:weight_fmlast_alex}
    
}
\subfigure[]{
    \includegraphics[width=\figurewidthtwo\columnwidth]{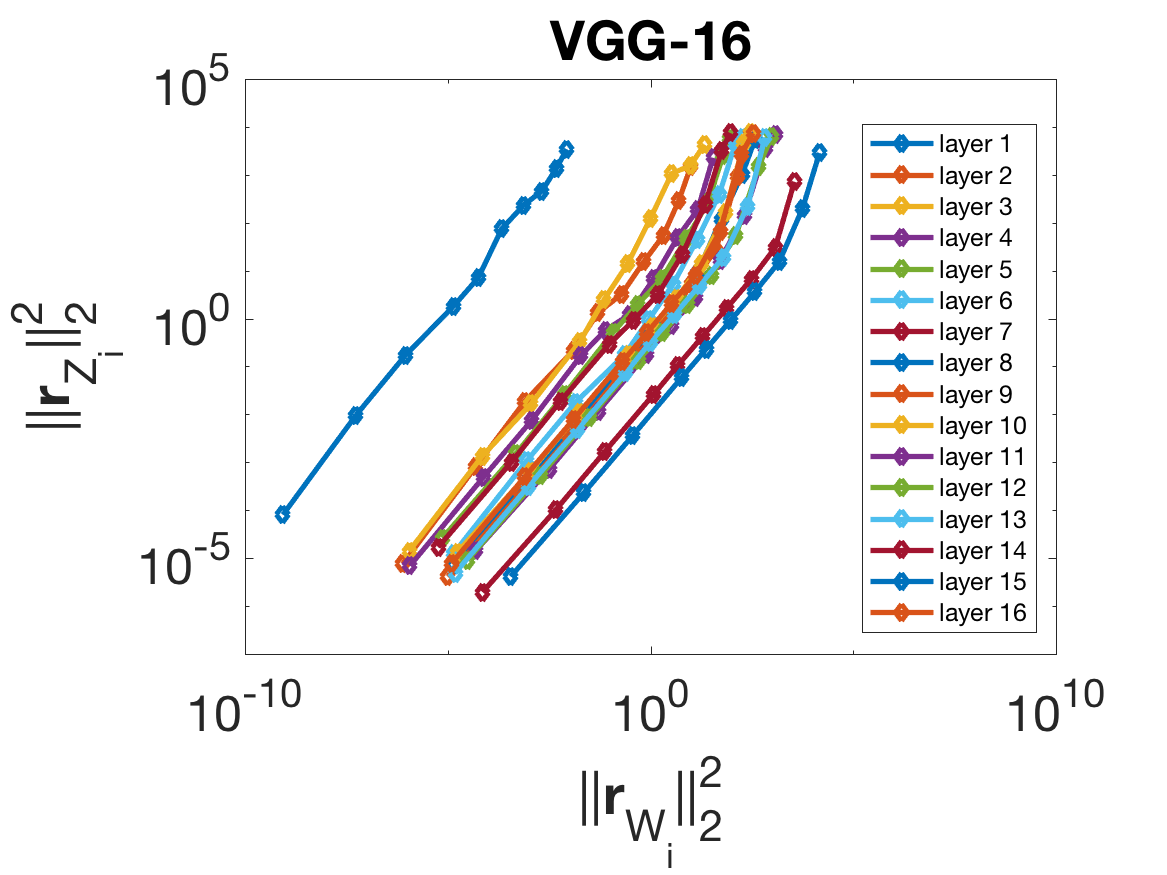}
    \label{fig:weight_fmlast_vgg}
    
}
\caption[]{The relationship between $||\textbf{r}_{W_i}||_2^2$ and $||\textbf{r}_{Z_i}||_2^2$.}
\label{fig:weight_fmlast}
\end{figure}

Fig.~\ref{fig:weight_fmlast} shows the relationship between the norm square of noise on quantized weight $||\textbf{r}_{W_i}||_2^2$ and $||\textbf{r}_{Z_i}||_2^2$ on different layers. When the quantization noise on weight is small, we can observe linear relationships. While it is interesting to see that, when the quantization noise is large, the curve does not follow exact linearity, and curves for earlier layers are not as linear as later layers. One possible explanation is that earlier layers in a DNN model are affected by more non-linear layers, such as ReLU and Max-pooling layers.
%
%
When the noise is large enough to reach the non-linear part of the layer functions (i.e. the zero point of the ReLU function), the curves become non-linear. It is worth noting that, when the non-linearity in most layers happens, the accuracy of the model is already heavily affected (become near zero). So this non-linearity would not affect our quantization optimization process.

\subsubsection{Additivity of measurements}
\label{sssec:additivity_of_measurements}

\begin{figure}[!t]
\centering
\subfigure[]{
    \includegraphics[width=\figurewidthtwo\columnwidth]{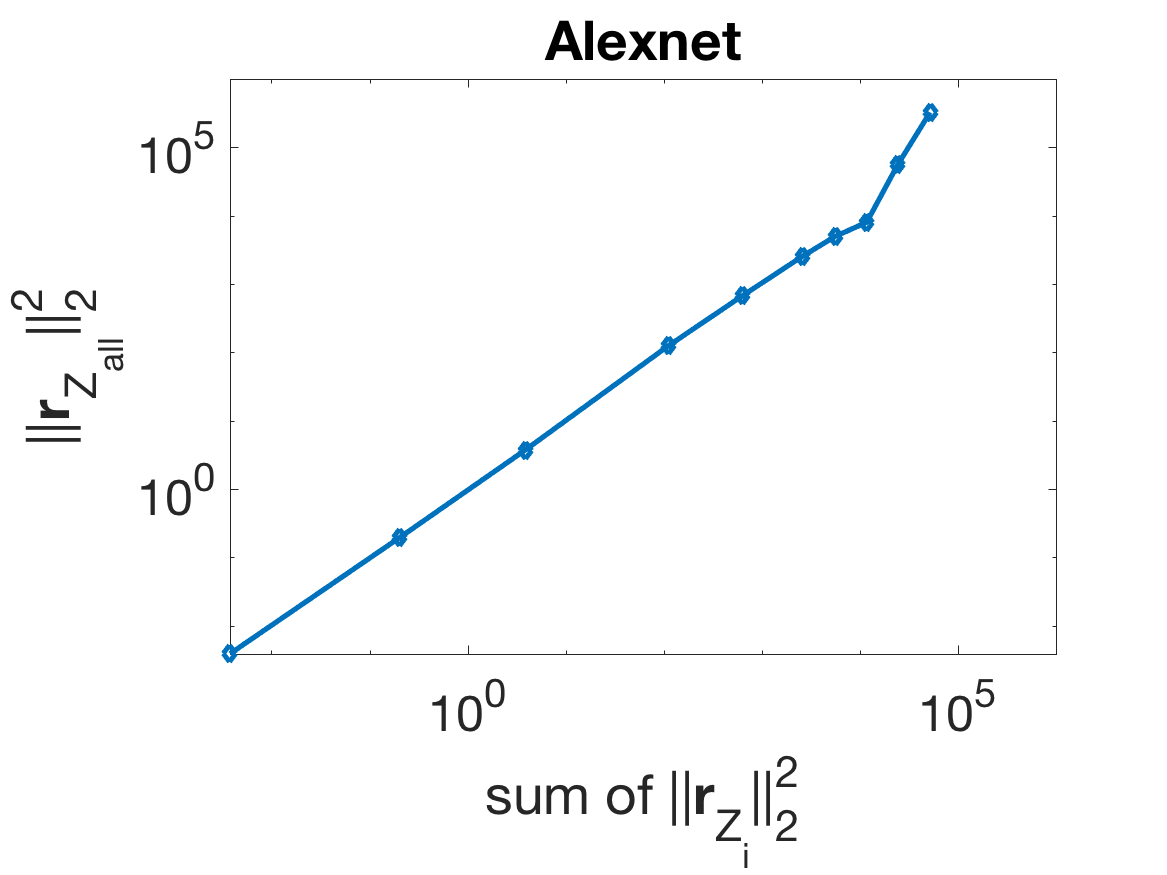}
    \label{fig:fmpred_fmtrue_alex}
    
}
\subfigure[]{
    \includegraphics[width=\figurewidthtwo\columnwidth]{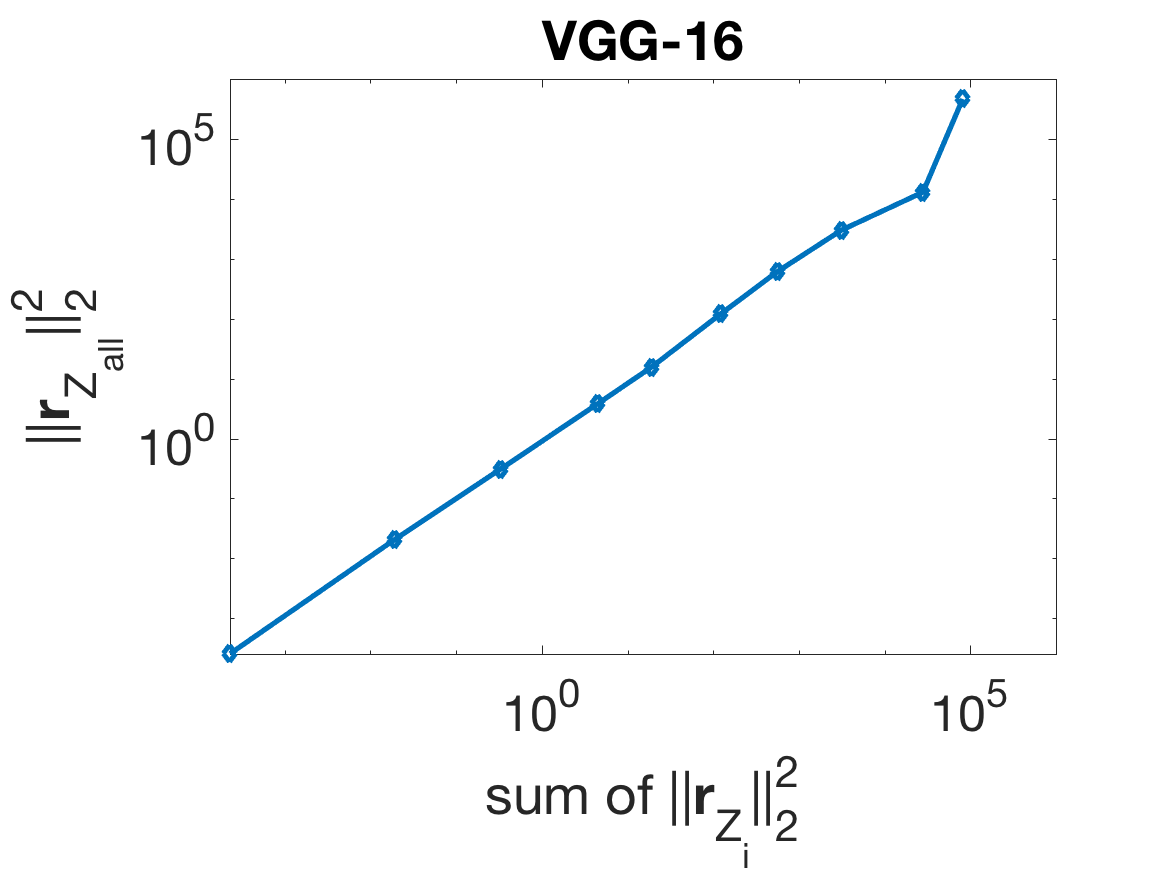}
    \label{fig:fmpred_fmtrue_vgg}
    
}
\caption[]{The value of $\sum_{i}^{N}||\textbf{r}_{Z_i}||_2^2$ when quantize each layer separately, compare to $||\textbf{r}_{Z}||_2^2$ when quantize all layers simultaneously.}
\label{fig:fmpred_fmtrue}
\end{figure}

Fig.~\ref{fig:fmpred_fmtrue} shows the relationship between $\sum_{i}^{N}||\textbf{r}_{Z_i|}|_2^2$ when we quantize each layer separately, and the value $||\textbf{r}_{Z}||_2^2$ when we quantize all layers together. We can see that when the quantization noise is small, the result closely follows our analysis that $||\textbf{r}_{Z}||_2^2=||\textbf{r}_{Z_i}||_2^2$; it validates the additivity of $||\textbf{r}_Z||_2^2$. When the quantization noise is large, the additivity of $||\textbf{r}_Z||_2^2$ is not accurate. This result fits our assumption in Eq.~(\ref{eq:convolution_value_noise_both_2}), where the additivity holds under the condition $||\textbf{r}_{W_i}||_2\ll ||W_i||_2$ for all layer $i$ in the DNN model.
When the noise is too high and we observe the inaccuracy of additivity, the model accuracy is already heavily degraded (near zero). Hence it does not affect the quantization optimization process which rather works in low noise regime.

\subsection{Optimal bit-width for models}
\label{ssec:optimal_bit-width_for_models}

After the validation of the proposed measurement, we conduct experiments to show the results on adaptive quantization. Here we use Alexnet~\cite{krizhevsky2012imagenet}, VGG-16~\cite{Simonyan14c}, GoogleNet~\cite{szegedy2015going}, and Resnet-50~\cite{he2016deep} to test our bit-width optimization approach. Similarly to the last experiments, the validation set of Imagenet is used. As the SQNR-based method~\cite{lin2016fixed} only works for convolutional layers, here we keep the fully connected layers with 16 bits
.

\begin{figure}[!t]
\centering
\subfigure[]{
    \includegraphics[width=\figurewidthtwo\columnwidth]{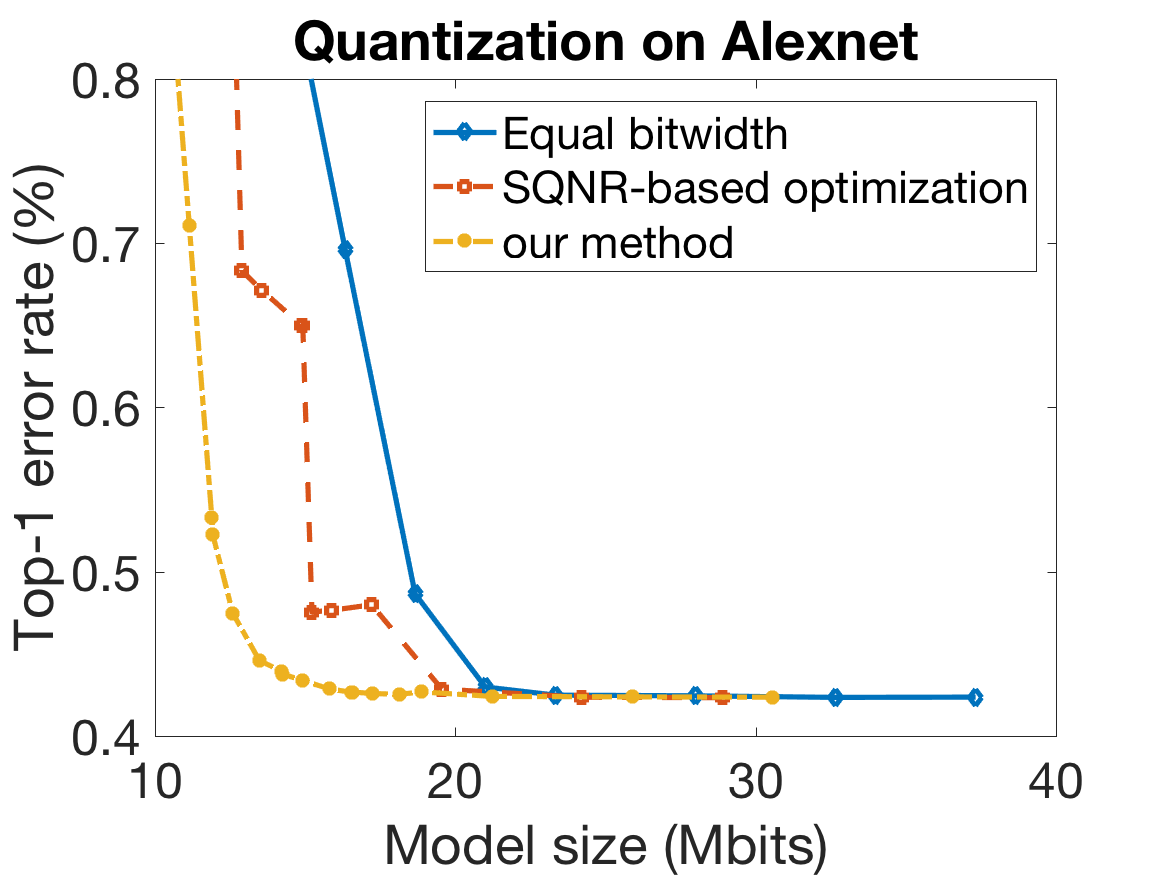}
    \label{fig:res_top1_compare_alex}
    
}
\subfigure[]{
    \includegraphics[width=\figurewidthtwo\columnwidth]{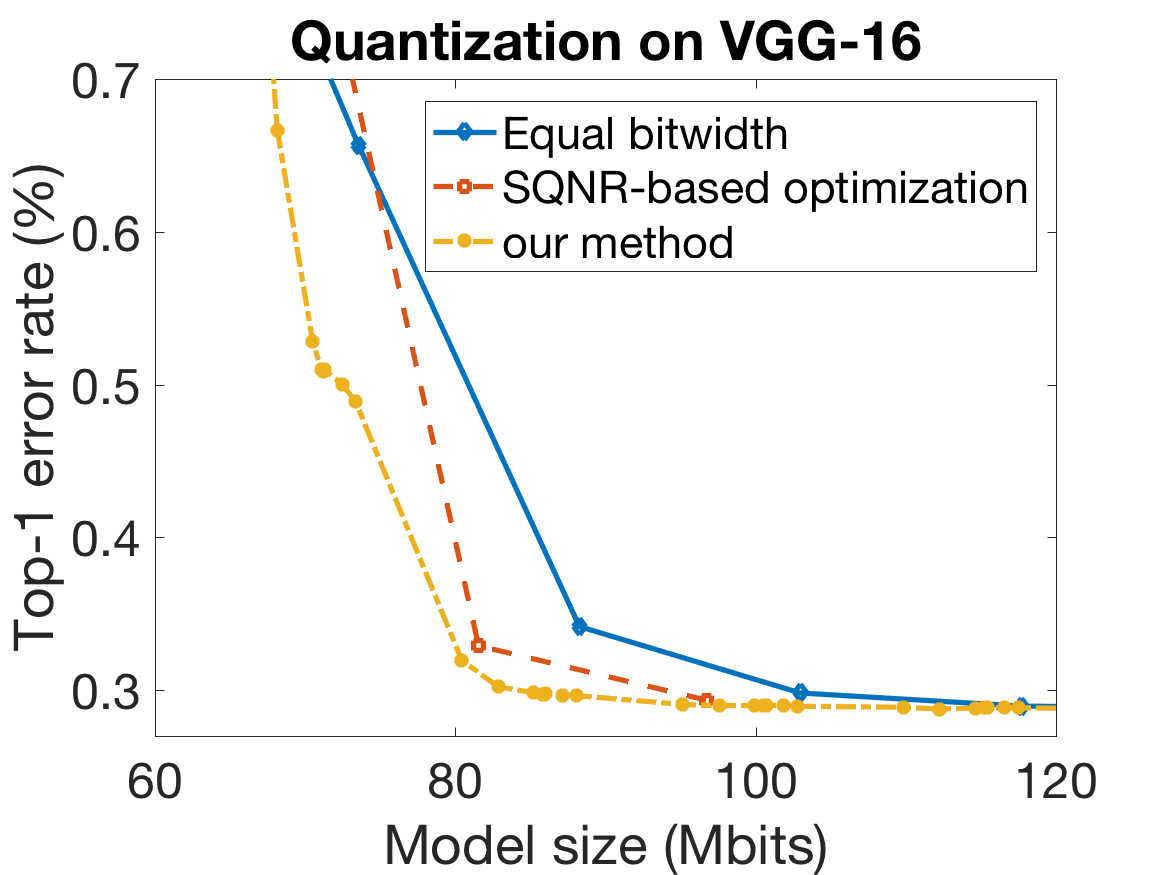}
    \label{fig:res_top1_compare_vgg}
    
}
\subfigure[]{
    \includegraphics[width=\figurewidthtwo\columnwidth]{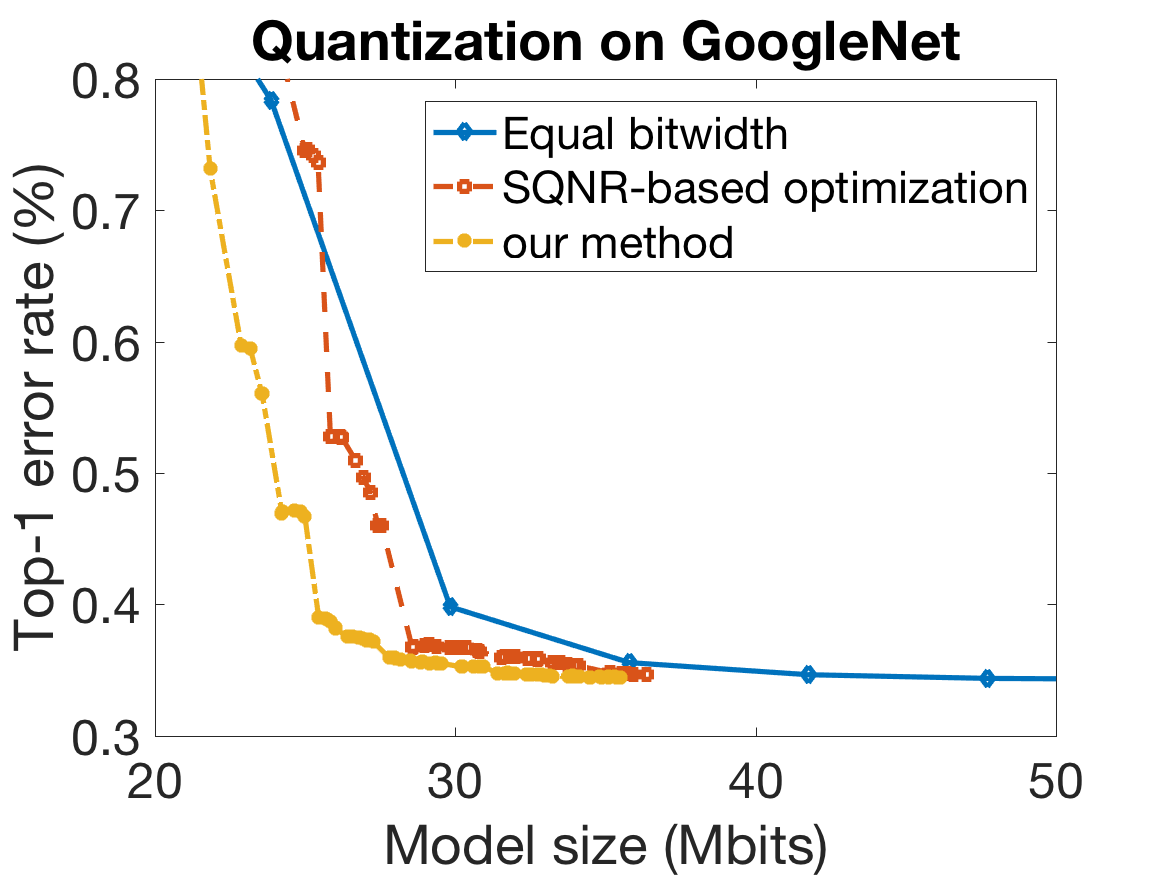}
    \label{fig:res_top1_compare_google}
    
}
\subfigure[]{
    \includegraphics[width=\figurewidthtwo\columnwidth]{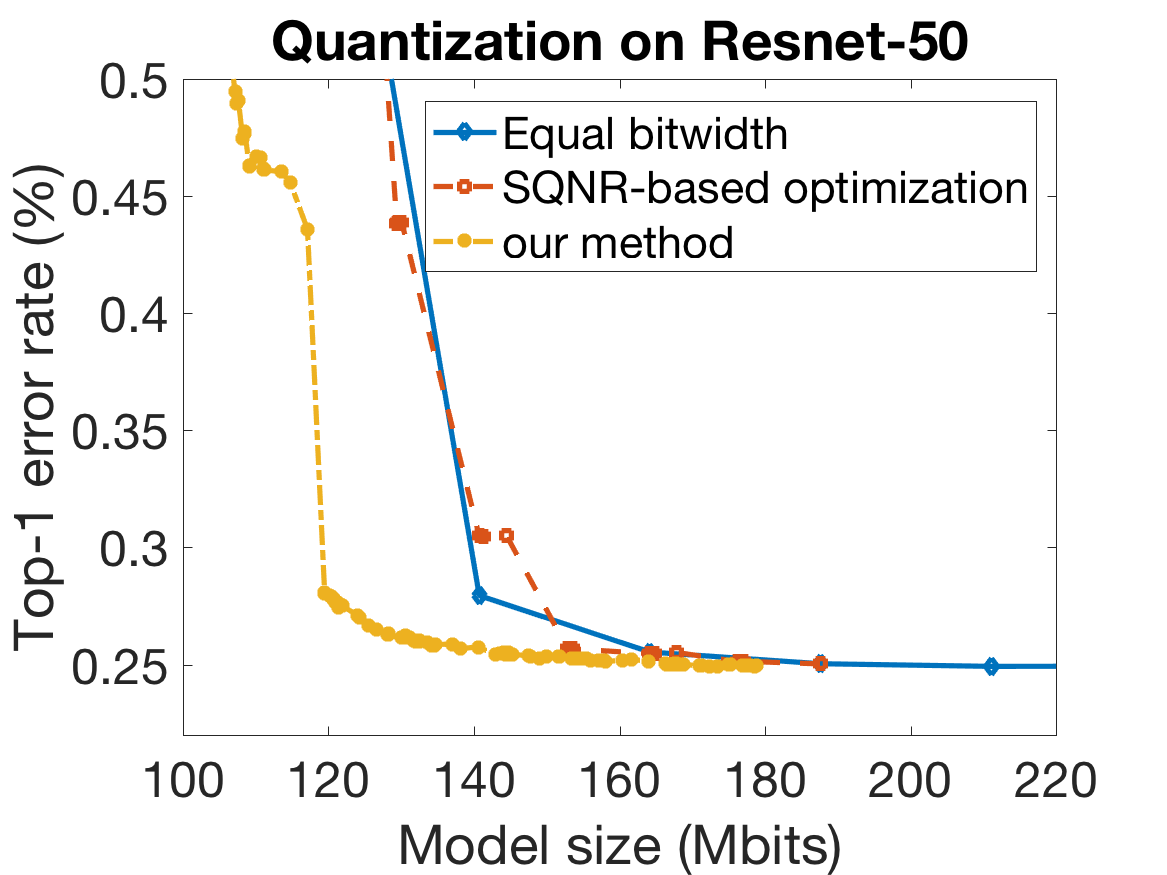}
    \label{fig:res_top1_compare_resnet}
    
}
\caption[]{Model size after quantization, v.s. accuracy. To compare with SQNR-based method~\cite{lin2016fixed}, only convolutional layers are quantized.}
\label{fig:res_top1_compare}
\end{figure}

Fig.~\ref{fig:res_top1_compare} shows the quantization results using our method, SQNR-based method~\cite{lin2016fixed}, and equal bit-width quantization. The equal bit-width quantization means that the number of quantization intervals in all layers are the same. For all three methods, we use uniform quantization for each layer. We can see that for all networks, our proposed method outperforms SQNR-based method, and achieves smaller model size for the same accuracy degradation. It is interesting to see that the SQNR-based method does not obviously outperform equal quantization on the Resnet-50 model. One possible reason is that Resnet-50 contains $1\times 1$ convolutional layers in its "bottleneck" structure, which is similar to fully connected layers. As the authors claim in~\cite{lin2016fixed}, the SQNR-based method does not work for fully connected layers.
Note that our method generates more datapoints on the figure, because the optimal bit-width for different layers may contain different decimals. And by rounding the optimal bit-width in different ways, we can generate more bit-width combinations than the SQNR-based methods.


The results for quantization on all layers are shown in Supplementary Material.
For Alexnet and VGG-16 model, our method achieves $30-40\%$ smaller model size with the same accuracy degradation, while for GoogleNet and Resnet-50, our method achieves $15-20\%$ smaller model size with the same accuracy degradation. These results indicate that our proposed quantization method works better for models with more diverse layer size and structures, like Alexnet and VGG.

%
\section{Conclusions} 
\label{sec:conclusions}

Parameter quantization is an important process to reduce the computation and memory costs of DNNs, and to deploy complex DNNs on mobile equipments. In this work, we propose an  efficient approach to optimize layer-wise bit-width for parameter quantization. We propose a method that relates quantization to model accuracy, and theoretically analyses this method. We show that the proposed approach is more general and accurate than previous quantization optimization approaches. Experimental results show that our method outperforms previous works, and achieves $20-40\%$ higher compression rate than SQNR-based methods and equal bit-width quantization. For future works, we will consider combining our method with fine-tuning and other model compression methods to achieve better model compression results.


\bibliographystyle{aaai}
\bibliography{refs}

\newpage
\input{supplementary}

\end{document}

%% file: supplementary.tex
%
\title{\textbf{Supplementary Material: Adaptive Quantization for Deep Neural Network}}
\author{\textbf{\Large
Yiren Zhou\textsuperscript{1}, Seyed-Mohsen Moosavi-Dezfooli\textsuperscript{2}, Ngai-Man Cheung\textsuperscript{1}, Pascal Frossard\textsuperscript{2}}\\
\textsuperscript{1}Singapore University of Technology and Design (SUTD)\\
\textsuperscript{2}École Polytechnique Fédérale de Lausanne (EPFL)\\
yiren\_zhou@mymail.sutd.edu.sg, ngaiman\_cheung@sutd.edu.sg\\
\{seyed.moosavi, pascal.frossard\}@epfl.ch\\
}

\maketitle

\section{Measuring the effect of quantization noise}

\subsection{Quantization noise}

Assume that conducting quantization on a value is equivalent to adding noise to the value:

\begin{equation}
w_q = w + r_w
\tag{\ref{eq:quantization_noise} revisited}
\end{equation}

Here $w$ is the original value, $w\in W$, $W$ is all weights in a layer. $w_q$ is the quantized value, and $r_w$ is the quantization noise. Assume we use a uniform quantizer, that the stepsize of the quantized interval is fixed. Then the quantization noise $r_w$ follows a uniform distribution in range $(-\frac{B}{2}, \frac{B}{2})$, where $B$ is the quantized interval. Based on this, $r_w$ has zero mean, and the variance of the noise $var(r_w)=\frac{B^2}{12}$. Then
we have
$E(r_w^2)=var(r_w)+E(r_w)^2=\frac{B^2}{12}$.

Follow the uniform quantization analysis in~\cite{you2010audio}, given weights $w\in W$ in a layer, $w\in (w_{min}, w_{max})$. If we quantize the weights by $b$ bits, the total number of interval would be $M=2^{b}$, and quantization interval would be $\frac{w_{min}-w_{max}}{2^{b}}$.
If we consider $\textbf{r}_w=(r_{w,1},\cdots,r_{w,N_W})$ as the quantization noise on all weights in $W$,
The expectation of noise square:

\begin{equation}
\begin{split}
E(||\textbf{r}_w||_2^2) &= \sum_{i}^{N_W}E(r_w^2) = N_W\frac{(w_{min}-w_{max})^2}{12}\cdot 4^{-{b}} \\
 &= p_w'\cdot e^{-\alpha\cdot {b}}
\end{split}
\tag{\ref{eq:quantization_noise_expectation} revisited}
\end{equation}

Where $p_w'=N_W\frac{(w_{min}-w_{max})^2}{12}$, $N_W$ is the number of weights in $W$, and $\alpha =ln(4)$. Eq.~(\ref{eq:quantization_noise_expectation}) indicates that every time we reduce the bit-width by 1 bit, $E(\textbf{r}_w^2)$ will increase by 4 times. This is equivalent to the quantization efficiency of 6dB/bit mentioned in~\cite{lin2016fixed}.

\subsection{The property of softmax classifier}
\label{ssec:the_property_of_softmax_classifier}

\begin{figure}[htbp]
\addtocounter{figure}{-1}
\renewcommand\thefigure{\ref{fig:model_simple}}
\includegraphics[width=\figurewidthone\columnwidth]{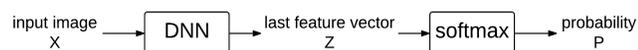}
\caption[]{Simple DNN model architecture.}
\end{figure}

Similar to the analysis in~\cite{pang2017robust}, we analyse the property of softmax classifier.

A DNN classifier can be expressed as a mapping function $\mathcal{F}(X,W):\mathscr{R}^d\to \mathscr{R}^L$, where $X\in \mathscr{R}^d$ is the input variable, $W$ is the parameters, and $L$ denotes the number of classes.

From Fig.~\ref{fig:model_simple}, here we divide the DNN into two parts. In the first part, we have a mapping function $\mathcal{G}(X,W):\mathscr{R}^d\to \mathscr{R}^L$, which maps input variables $X\in \mathscr{R}^d$ into the feature vectors $Z\in \mathscr{R}^L$ for the last layer of DNN. In the second part, we have the softmax function $softmax(\textbf{z}):\mathscr{R}^L\to \mathscr{R}^L$ as $softmax(z_i)=exp(z_i)/\sum^{L}_{i=1}exp(z_i)$, $i\in [L]$, where $[L]:={1, ..., L}$.

The final classification result can be calculated by picking the maximum value of the softmax value: $\argmax_i softmax(z_i)$, $i\in [L]$. Note that this is equivalent to picking the maximum value for feature vector $\textbf{z}$: $\argmax_i z_i$, $i\in [L]$. So we can see that the softmax classifier has a
linear decision boundary in the feature vectors $Z$~\cite{pang2017robust}.

\subsection{Proof of Lemma~\ref{lem:linear}}
\label{ssec:proof_of_lemma}

\primelemma*

\proof
Based on Theorem 1 in \cite{NIPS2016_6331}, for an $L$-class classifier, the norm of a random noise to fool the classifier can be bounded from below by
\begin{equation}
    d\|\textbf{r}^*\|^2_2\leq\beta(\delta')\|\textbf{r}_Z\|^2_2,
\end{equation}
with a probability exceeding $1-2(L+1)\delta'$, where $\beta(\delta')=1+2\sqrt{\ln(1/\delta')}+2\ln(1/\delta')$. For the softmax layer, $L=d-1$, therefore one can write
\begin{equation}
    \mathbb{P}\left(d\|\textbf{r}^*\|^2_2\leq\beta(\delta')\|\textbf{r}_Z\|^2_2\right)\geq 1-2d\delta'.
\end{equation}
Furthermore, $\|\textbf{r}^*\|_2^2=(z_{(1)}-z_{(2)})^2/2$, where $z_{(i)}$ is $i^{th}$ largest element of $z$. Put $\delta'=\delta/d$, hence
\begin{equation}
    \mathbb{P}\left(d\|\textbf{r}^*\|^2_2\leq\beta(\delta/d)\|\textbf{r}_Z\|^2_2\right)\geq 1-2\delta.
\end{equation}
From the other hand,
\begin{equation}
    \begin{split}
        \beta(\delta/d)&=1+2\sqrt{\ln(d/\delta)}+2\ln(d/\delta)
        \\ &\geq 1+4\ln(d/\delta)
        \\ &\geq \ln{d}\left(5+4\ln(1/\delta)\right)=\gamma(\delta)\ln{d}.
    \end{split}
\end{equation}
Therefore,
\begin{equation}
    \begin{split}
        \mathbb{P}\left(\frac{d}{\ln{d}}\|\textbf{r}^*\|^2_2\leq\gamma(\delta)\|\textbf{r}_Z\|^2_2\right)&\geq\mathbb{P}\left(d\|\textbf{r}^*\|^2_2\leq\beta(\delta/d)\|\textbf{r}_Z\|^2_2\right)
        \\&\geq 1-2\delta,
    \end{split}
    \label{eq:proof_final}
\end{equation}
which concludes the proof.
\endproof

\subsection{Relationship between accuracy and noise}
\label{ssec:relationship_between_accuracy_and_noise}

The original quantization optimization problem:

\begin{equation}
\begin{split}
&~min \sum_{i=1}^{N}s_i\cdot b_i \\
s.t.~&~acc_{\mathcal{F}} - acc_{\mathcal{F'}} \leq \Delta_{acc}
\end{split}
\tag{\ref{eq:optimization_0} revisited}
\end{equation}

Lemma~\ref{lem:linear} states that if the norm of random noise is $o\left((z_{(1)}-z_{(2)})\sqrt{d/\ln d}\right)$, it does not change the classifier decision with high probability. In particular, from Lemma~\ref{lem:linear} (Eq.~(\ref{eq:proof_final}) in specific), the probability of misclassification can be expressed as:

\begin{equation}
\mathbb{P}( \|\textbf{r}_Z\|_2^2 \leq \frac{d}{\gamma(\delta)\ln{d}}\frac{(z_{(1)}-z_{(2)})^2)}{2}) \leq 2\delta
\label{eq:prob_acc}
\end{equation}

Eq.~(\ref{eq:prob_acc}) suggest that as we limit the noise to be less than $\frac{d}{\gamma(\delta)\ln{d}}\frac{(z_{(1)}-z_{(2)})^2}{2}$, the probability of misclassification should be less than $2\delta$.

Based on Lemma~\ref{lem:linear} and Eq.~(\ref{eq:prob_acc}), we formulate the relationship between the noise and model accuracy. Assume that we have a model $\mathcal{F}$ with accuracy $acc_{\mathcal{F}}$. After adding random noise $\textbf{r}_Z$ on the last feature map $Z$, the model accuracy drops $\Delta_{acc}$. If we assume that the accuracy degradation is caused by the noise $\textbf{r}_Z$, we can see that the $\delta$ value in Eq.~(\ref{eq:prob_acc}) is closely related to $\Delta_{acc}$:

\begin{equation}
\begin{split}
& 2\delta \sim \frac{\Delta_{acc}}{acc_{\mathcal{F}}} \\
\Rightarrow & \gamma(\delta) \sim \gamma(\frac{\Delta_{acc}}{2acc_{\mathcal{F}}})
\end{split}
\label{eq:prob_error}
\end{equation}

If we have:

\begin{equation}
\theta(\Delta_{acc}) = \frac{d}{\gamma(\frac{\Delta_{acc}}{2acc_{\mathcal{F}}})\ln{d}}
\tag{\ref{eq:theta_define} revisited}
\end{equation}

From Eq.~(\ref{eq:prob_acc}), we have:

\begin{equation}
\begin{split}
\mathbb{P}( \|\textbf{r}_Z\|_2^2 < \theta(\Delta_{acc})\frac{(z_{(1)}-z_{(2)})^2}{2} ) \leq 2\delta \sim \frac{\Delta_{acc}}{acc_{\mathcal{F}}}
\end{split}
\label{eq:norm_square_error}
\end{equation}

Eq.~(\ref{eq:norm_square_error}) indicates that by limiting noise $r_z$ to be less than $\theta(\Delta_{acc})\frac{(z_{(1)}-z_{(2)})^2}{2}$, we can approximately assume that model accuracy drops less than $\Delta_{acc}$. As $\gamma(\delta)$ is strictly decreasing, we can see that $\theta(\Delta_{acc})$ is strictly increasing w.r.t. $\Delta_{acc}$. So as the model has higher accuracy degradation $\Delta_{acc}$, the noise limitation also increase.

Based on Eq.~(\ref{eq:norm_square_error}),
we have the relation between accuracy degradation and noise $\textbf{r}_Z$ as:

\begin{equation}
\begin{split}
&~acc_{\mathcal{F}} - acc_{\mathcal{F'}} \leq \Delta_{acc} \Rightarrow \\
&~\|\textbf{r}_Z\|_2^2 < \theta(\Delta_{acc})\frac{(z_{(1)}-z_{(2)})^2}{2}
\end{split}
\tag{\ref{eq:relation_acc_noise} revisited}
\end{equation}


\subsection{Calculation of noise on convolutional layer} 
\label{ssec:calculation_of_noise_on_convolutional_layer}

Given a convolutional layer input $A$ with size $M\times N\times C$, conv kernel $K$ with size $M_k\times N_k\times C\times D$, and stride size $s$, we have the output feature map $Z$ with size $(M/s)\times (N/s)\times D$. Here $M$ and $N$ are the height and width of input, $C$ is the number of channel of input. $M_k$ and $N_k$ are the height and width of conv kernel, $D$ is the depth of output feature map.

The analysis on fully connected layers will be similar to the analysis on convolutional layers. It can be considered as a special case of convolutional layers when $M$, $N$, $M_k$, and $N_k$ are equal to 1.

Based on the definition of convolutional operation, for a single value $z\in Z$, the value is calculated as:

\begin{equation}
\begin{split}
z_{p, q, d} & = \sum_{m_k=1}^{M_k}\sum_{n_k=1}^{N_k}\sum_{c=1}^{C}a_{m_k+p\cdot s, n_k+q\cdot s, c}\cdot w_{m_k, n_k, c, d} + b_d \\
            & = \sum_{i=1}^{M_k\cdot N_k\cdot C}a_i\cdot w_{i, d} + b_d
\end{split}
\label{eq:convolution_value}
\end{equation}

where $w$ is the weight, and $b$ is the bias. $p\in \{1, ..., M/s\}$ and $q\in \{1, ..., N/s\}$.

As we consider noise on both input feature maps and weights, the Eq.~(\ref{eq:convolution_value}) will become:

\begin{equation}
\begin{split}
z_{p, q, d} & = (\sum_{i=1}^{M_k\cdot N_k\cdot C}a_i\cdot w_{i, d} + b_d) \\
            & + (\sum_{i=1}^{M_k\cdot N_k\cdot C}(a_i\cdot r_{w_{i, d}}
            + r_{a_i}\cdot w_{i, d} + r_{a_i}\cdot r_{w_{i, d}}) + r_{b_d})
\end{split}
\label{eq:convolution_value_noise_both}
\end{equation}

Then the noise term of $z_{p, q, d}$ can be expressed as:

\begin{equation}
\begin{split}
r_{z_{p, q, d}} & = \sum_{i=1}^{M_k\cdot N_k\cdot C}(a_i\cdot r_{w_{i, d}} + r_{a_i}\cdot w_{i, d} + r_{a_i}\cdot r_{w_{i, d}}) + r_{b_d} \\
                        & \approx \sum_{i=1}^{M_k\cdot N_k\cdot C}(a_i\cdot r_{w_{i, d}} + r_{a_i}\cdot w_{i, d})
\end{split}
\tag{\ref{eq:convolution_value_noise_both_2} revisited}
\end{equation}

Note that the term $r_{b_d}$ can be ignored under the assumption that $w$ and $b$ have same bit-width quantization. The term $r_{a_i}\cdot r_{w_{i, d}}$ can be ignored under the assumption that $||r_a||_2\ll ||a||_2$ and $||r_w||_2\ll ||w||_2$.

\subsection{Additivity of $\|\textbf{r}_Z\|_2^2$}
\label{ssec:additivity_of_r_z}

\begin{figure}[htbp]
\addtocounter{figure}{-1}
\renewcommand\thefigure{\ref{fig:model_layer}}
\subfigure[]{\includegraphics[width=\figurewidthone\columnwidth]{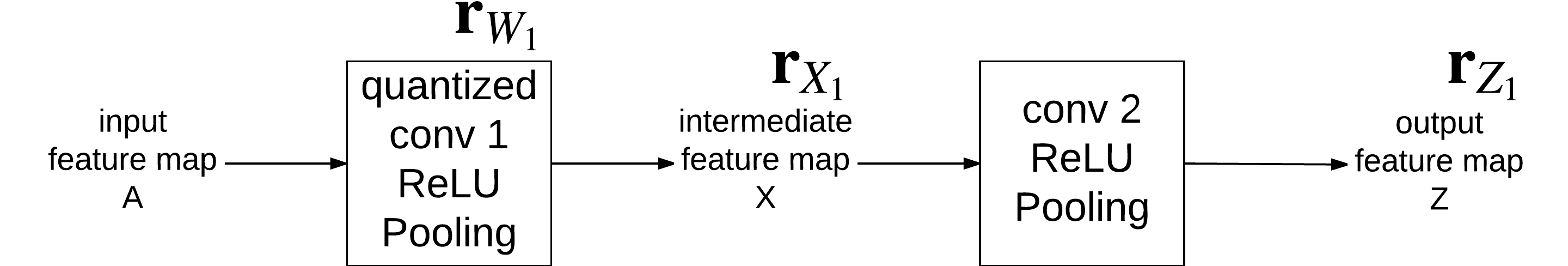}
}
\subfigure[]{\includegraphics[width=\figurewidthone\columnwidth]{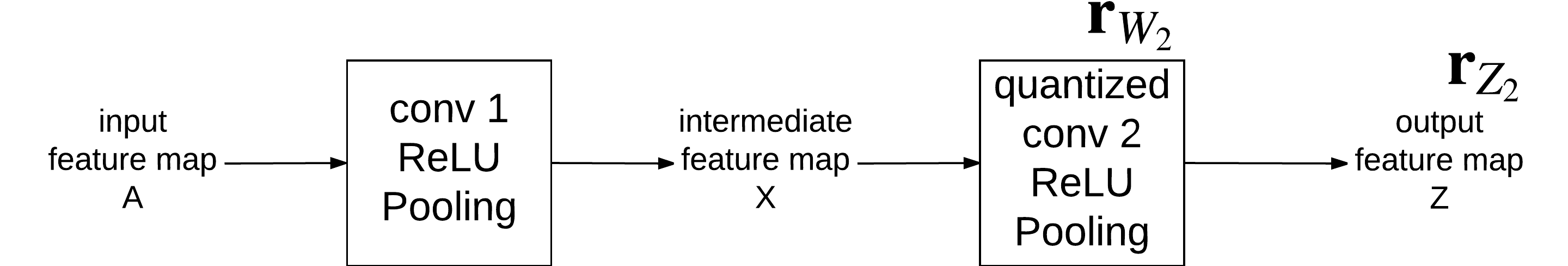}
}
\subfigure[]{\includegraphics[width=\figurewidthone\columnwidth]{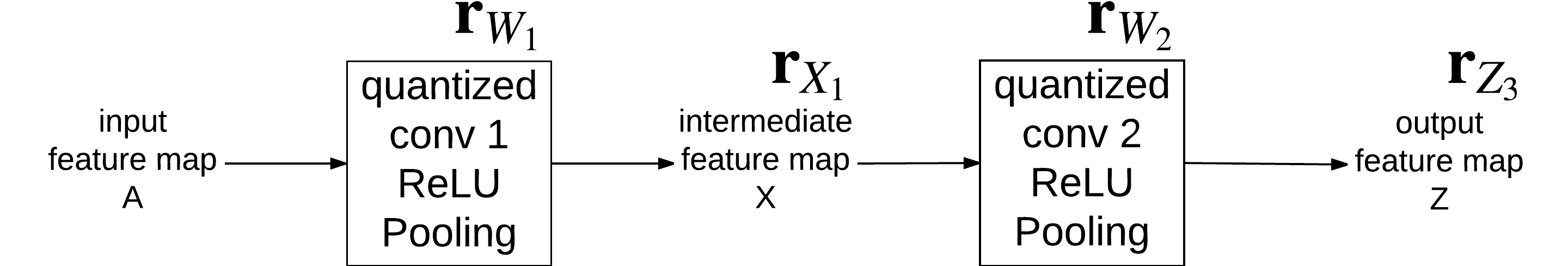}
}
\caption[]{Effect of adding noise to multiple layers.}
\end{figure}

Fig.~\ref{fig:model_layer} shows a 2-layer module inside a DNN model. Given input feature map $A$, after the first conv layer, an intermediate feature map $X$ is generated, then after the second conv layer, output feature map $Z$ is generated.

In Fig.~\ref{sfig:model_layer_1}, the weight of the first conv layer $W_1$ is quantized, result in the quantization noise $\textbf{r}_{W_1}$ on weight $W_1$, then the noise passes to feature map $X$ with noise $\textbf{r}_{X_1}$, and feature map $Z$ with $\textbf{r}_{Z_1}$.

In Fig.~\ref{sfig:model_layer_2}, the weight of the second conv layer $W_2$ is quantized, result in the quantization noise $\textbf{r}_{W_2}$ on weight $W_2$, then the noise passes to feature map $Z$ with $\textbf{r}_{Z_2}$.

In Fig.~\ref{sfig:model_layer_3}, we quantize the weight of both layer $W_1$ and $W_2$, with quantization noise $\textbf{r}_{W_1}$ and $\textbf{r}_{W_2}$, the noise passes to $Z$ with noise $\textbf{r}_{Z_3}$.
Based on the discussion in previous section, $\textbf{r}_{Z_3}\approx \textbf{r}_{Z_1}+\textbf{r}_{Z_2}$.
Given a particular $\textbf{z}\in Z$, we have $\textbf{z}=(z_1,\cdots,z_L)$, where $\textbf{z}\in \mathbb{R}^L$.
Then $\textbf{r}_{\textbf{z}_1}=(r_{z_1,1},\cdots,r_{z_1,L})$, $\textbf{r}_{\textbf{z}_2}=(r_{z_2,1},\cdots,r_{z_2,L})$, and $\textbf{r}_{\textbf{z}_3}\approx(r_{z_1,1}+r_{z_2,1},\cdots,r_{z_1,L}+r_{z_2,L})$. $||\textbf{r}_{\textbf{z}_3}||_2^2$ can be calculated as:

\begin{equation}
\begin{split}
||\textbf{r}_{\textbf{z}_3}||_2^2
& = \sum_{i=1}^{L}r_{z_3,i}^2
\\
& \approx \sum_{i=1}^{L}(r_{z_1,i} + r_{z_2,i})^2 \\
& 
= \sum_{i=1}^{L}r_{z_1,i}^2 + \sum_{i=1}^{L}r_{z_2,i}^2 + 2\cdot\sum_{i=1}^{L}r_{z_1,i}\cdot r_{z_2,i} \\
& = ||\textbf{r}_{\textbf{z}_1}||_2^2 + ||\textbf{r}_{\textbf{z}_2}||_2^2 + 2\cdot\sum_{i=1}^{L}r_{z_1,i}\cdot r_{z_2,i} \\
& \doteq ||\textbf{r}_{\textbf{z}_1}||_2^2 + ||\textbf{r}_{\textbf{z}_2}||_2^2
\end{split}
\tag{\ref{eq:two_layer_add} revisited}
\end{equation}

The last equality holds under the assumption that $\textbf{r}_{\textbf{z}_1}$ and $\textbf{r}_{\textbf{z}_2}$ are independent. Which is reasonable in our case, as $\textbf{r}_{\textbf{z}_1}$ and $\textbf{r}_{\textbf{z}_2}$ are caused by $\textbf{r}_{W_1}$ and $\textbf{r}_{W_2}$ which are two independent quantization noise.

\section{Layer-wise bit-width optimization}

\subsection{Solving the optimization problem}
\label{ssec:solving_the_optimization_problem}

Consider the optimization problem:

\begin{equation}
\begin{split}
&~min \sum_{i=1}^{N}s_i\cdot b_i \\
s.t.& \sum_{i=1}^{N}\frac{p_i}{t_i}\cdot e^{-\alpha \cdot b_i}\leq C
\end{split}
\tag{\ref{eq:optimization_1} revisited}
\end{equation}

The form can be written as a dual problem:

\begin{equation}
\begin{split}
min \sum_{i=1}^{N}\frac{p_i}{t_i}\cdot e^{-\alpha \cdot b_i} \\
s.t. \sum_{i=1}^{N}s_i\cdot b_i=C'
\end{split}
\label{eq:optimization}
\end{equation}

We convert this to Lagrange function:

\begin{equation}
min \sum_{i=1}^{N}\frac{p_i}{t_i}\cdot e^{-\alpha \cdot b_i} - \lambda \cdot \sum_{i=1}^{N}s_i\cdot b_i
\label{eq:optimization_lagrange}
\end{equation}

Using KKT conditions, we have that the optimal value can be reached when:

\begin{equation}
\frac{p_1\cdot e^{-\alpha \cdot b_1}}{t_1 \cdot s_1}=\frac{p_2\cdot e^{-\alpha \cdot b_2}}{t_2 \cdot s_2}=\cdots=\frac{p_N\cdot e^{-\alpha \cdot b_N}}{t_N \cdot s_N}
\tag{\ref{eq:optimization_result} revisited}
\end{equation}

\subsection{Algorithm to calculate optimal bit-width}
\label{ssec:algorithm_to_calculate_optimal_bit_width}

The algorithms to calculate $t_i$, $p_i$, and $b_i$ is shown in Alg.~\ref{alg:calculate_t_i},~\ref{alg:calculate_p_i}, and~\ref{alg:calculate_b_i}, respectively.

\begin{algorithm}
\caption{Calculate $t_i$}
\label{alg:calculate_t_i}
\begin{algorithmic}[1]
\Procedure{cal\_t}{}

\textbf{Input:}

Dataset $\mathcal{D}$; label $\mathcal{L}$; accuracy degradation $\Delta_{acc}$;

model $\mathcal{F}$; model before softmax: $\mathcal{G}$;

weights $W_i\in W$ in layer $i$;

$mean_{\textbf{r}^*} \gets \frac{1}{|\mathcal{D}|}\sum_{\textbf{x}\in \mathcal{D}}\frac{(z_{(1)}-z_{(2)})^2}{2}$;

$acc_{\mathcal{F}} \gets \frac{|\mathcal{F}(\mathcal{D},W)=\mathcal{L}|}{|\mathcal{D}|}$.

\textbf{Output:}

$t_i, i=\{1,\cdots,N\}$

\For {layer $i = 1\to N$}
    \State $\textbf{r}_{W_i}^{'}=\mathcal{U}(-0.5, 0.5)$
    \State $k=k_{min}=10^{-5}, k_{max}=10^3$
    \While {$acc_{\mathcal{F}}-acc_{\mathcal{F'}}\neq\Delta_{acc}$}
        \If {$acc_{\mathcal{F}}-acc_{\mathcal{F'}}<\Delta_{acc}$}
            $k_{min} \gets k$
        \Else {$k_{max} \gets k$}
        \EndIf
        \State $k=\sqrt{k_{min}\cdot k_{max}}$
        \State $\textbf{r}_{W_i} \gets k\cdot \textbf{r}_{W_i}^{'}$
        \State $acc_{\mathcal{F'}} \gets \frac{|\mathcal{F}(\mathcal{D},W_i + \textbf{r}_{W_i})=\mathcal{L}|}{|\mathcal{D}|}$
    \EndWhile
    \State $\textbf{r}_{\textbf{z}_i}=\mathcal{G}(\textbf{x},W_i)-\mathcal{G}(\textbf{x},W_i+\textbf{r}_{W_i})$
    \State $mean_{\textbf{r}_{\textbf{z}_i}}=\frac{1}{|\mathcal{D}|}\sum_{\textbf{x}\in \mathcal{D}}||\textbf{r}_{\textbf{z}_i}||_2^2$
    \State $t_i = \frac{mean_{\textbf{r}_{\textbf{z}_i}}}{mean_{\textbf{r}^*}}$
\EndFor
\Return $t_i, i=\{1,\cdots,N\}$
\EndProcedure
\end{algorithmic}
\end{algorithm}

\begin{algorithm}
\caption{Calculate $p_i$}
\label{alg:calculate_p_i}
\begin{algorithmic}[1]
\Procedure{cal\_p}{}

\textbf{Input:}

Dataset $\mathcal{D}$; $\alpha=ln(4)$;

model before softmax: $\mathcal{G}$;

weights $W_i\in W$ in layer $i$;

quantization bit-width $b_i$ for layer $i$.

\textbf{Output:}

$p_i, i=\{1,\cdots,N\}$

\For {layer $i = 1\to N$}
    \State $W_{i,q}=quantize(W_i,b_i)$
    \State $\textbf{r}_{\textbf{z}_i}=\mathcal{G}(\textbf{x},W_i)-\mathcal{G}(\textbf{x},W_{i,q})$
    \State $mean_{\textbf{r}_{\textbf{z}_i}}=\frac{1}{|\mathcal{D}|}\sum_{\textbf{x}\in \mathcal{D}}||\textbf{r}_{\textbf{z}_i}||_2^2$
    \State $p_i=\frac{mean_{\textbf{r}_{\textbf{z}_i}}}{e^{-\alpha\cdot b_i}}$
\EndFor
\Return $p_i, i=\{1,\cdots,N\}$
\EndProcedure
\end{algorithmic}
\end{algorithm}

\begin{algorithm}
\caption{Calculate $b_i$}
\label{alg:calculate_b_i}
\begin{algorithmic}[1]
\Procedure{cal\_b}{}

\textbf{Input:}

Parameter $t_i$, $p_i$, $i=\{1,\cdots,N\}$; $\alpha=ln(4)$;

size $s_i$ of layer $i$;

quantization bit-width $b_1$ for layer $1$.

\textbf{Output:}

$b_i, i=\{2,\cdots,N\}$

\For {layer $i = 2\to N$}
    \State $b_i=b_1+\frac{1}{\alpha}\cdot ln(\frac{p_i\cdot t_1\cdot s_1}{p_1\cdot t_i\cdot s_i})$
\EndFor
\Return $b_i, i=\{2,\cdots,N\}$
\EndProcedure
\end{algorithmic}
\end{algorithm}

\subsection{Comparison with SQNR-based approach} 
\label{ssec:comparison_with_sqnr_based_approach}

From our work, the optimal quantization bit-width for each layer is reached when Eq.~(\ref{eq:optimization_result}) is fulfilled.

From the paper we have that $||r_{Z_i}||_2^2=p_i\cdot e^{-\alpha\cdot b_i}$ is the noise norm on last feature map $Z$ when we apply $b_i$-bit quantization on layer $i$, $s_i$ is the size of layer $i$, and $t_i$ is the robustness parameter for layer $i$.

Now we focus on the optimization result for SQNR-based method~\cite{lin2016fixed}. For the SQNR-based method, the optimal quantization bit-width is achieved when

\begin{equation}
\beta_i - \beta_j = \frac{10log(\rho_j/\rho_i)}{\kappa},
\label{eq:sqnr_result}
\end{equation}

where $\beta_i$ is the bit-width for layer $i$, $\rho_i$ is the number of parameter in layer $i$. $\kappa$ is the quantization efficiency, which corresponds to the $\alpha$ parameter in our work.

If we rewrite Eq.~(\ref{eq:sqnr_result}) using our notation, the optimization result for SQNR-based method becomes

\begin{equation}
\begin{split}
&~b_i - b_j = \frac{10log(s_j/s_i)}{\alpha'} \\
\Rightarrow &~\frac{\alpha'}{10}b_i - \frac{\alpha'}{10}b_j = log(\frac{s_j}{s_i}) \\
\Rightarrow &~\frac{e^{\frac{\alpha'}{10}b_i}}{e^{\frac{\alpha'}{10}b_j}} = \frac{s_j}{s_i} \\
\Rightarrow &~\frac{e^{-\frac{\alpha'}{10}b_i}}{s_i} = \frac{e^{-\frac{\alpha'}{10}b_j}}{s_j}
\end{split}
\label{eq:sqnr_modify}
\end{equation}

Let $\alpha=\alpha'/10$, we have the optimization result for SQNR-based method as

\begin{equation}
\frac{e^{-\alpha \cdot b_1}}{s_1}=\frac{e^{-\alpha \cdot b_2}}{s_2}=\cdots=\frac{e^{-\alpha \cdot b_N}}{s_N}
\tag{\ref{eq:optimization_result_SQNR} revisited}
\end{equation}

Note that compared with our result in Eq.~(\ref{eq:optimization_result}), the parameters $p_i$ and $t_i$ are missing. This is consistent with the assumption of the SQNR-based approach, that if two layers have the same bit-width for quantization, they would have the same SQNR value, thus the affection on accuracy are equal. This makes the SQNR-based approach a special case of our approach, when all layers in the DNN model have the equal affection on model accuracy under the same bit-width.

\section{Experimental results}

\begin{figure}[!t]
\centering
\subfigure{
    \includegraphics[width=\figurewidthtwo\columnwidth]{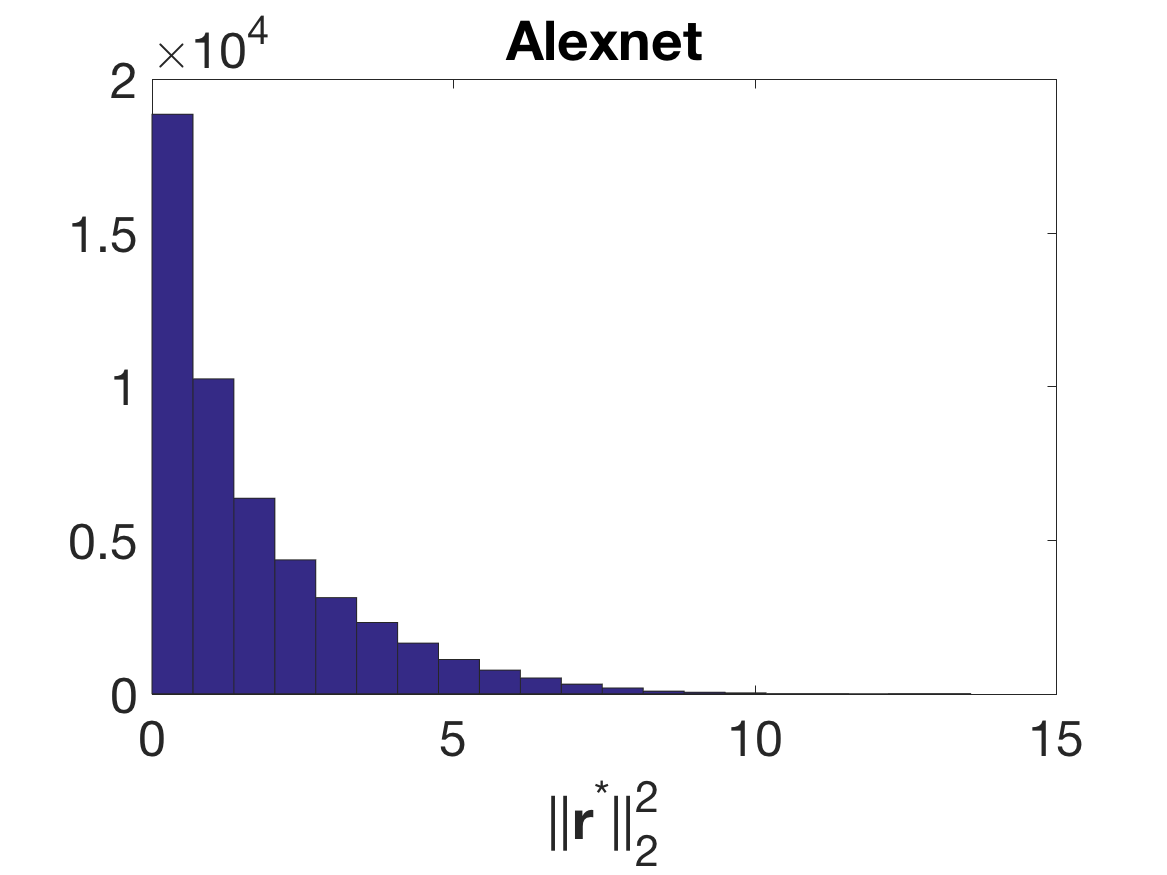}
}
\caption[]{Histogram of $||\textbf{r}^*||_2^2$ value for Alexnet on Imagenet validation set.}
\label{fig:adversarial_stat}
\end{figure}

\begin{figure}[!t]
\centering
\subfigure[]{
    \includegraphics[width=\figurewidthtwo\columnwidth]{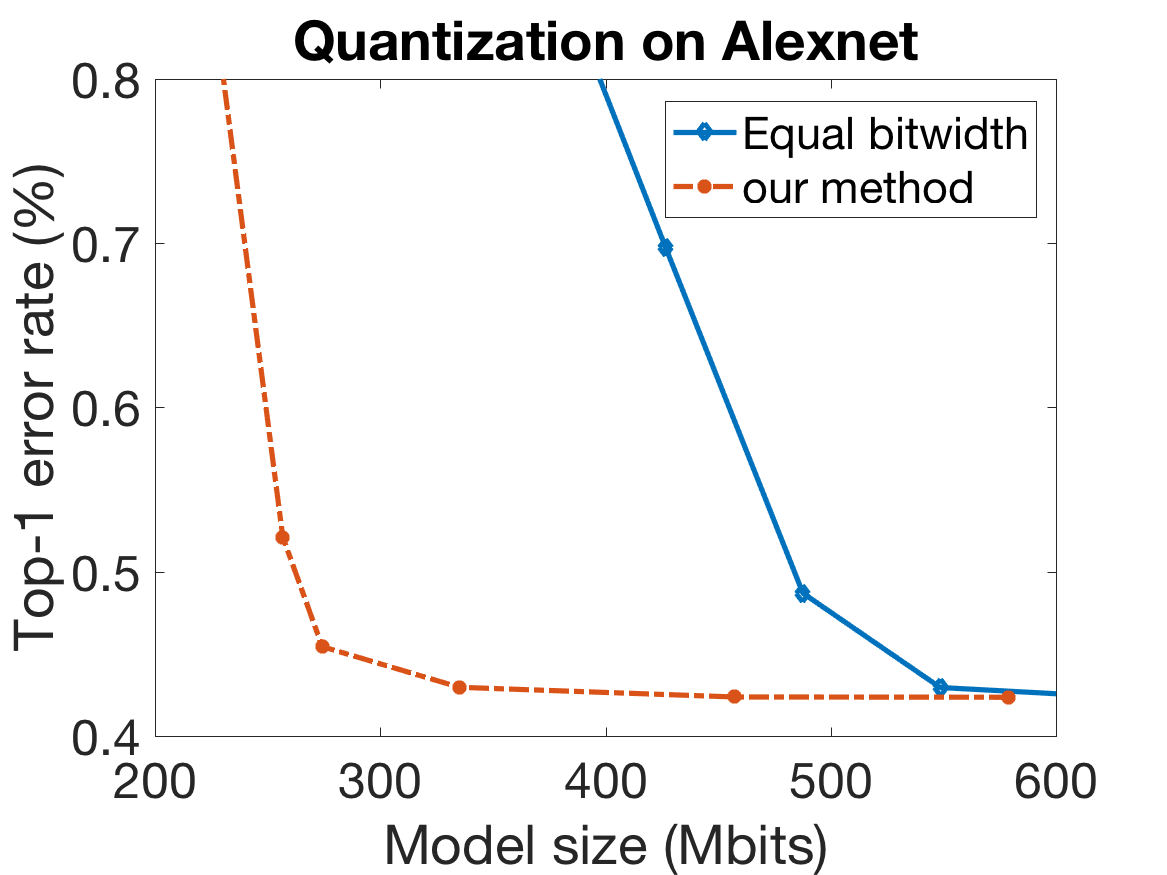}
    \label{fig:res_top1_compare_all_alex}
}
\subfigure[]{
    \includegraphics[width=\figurewidthtwo\columnwidth]{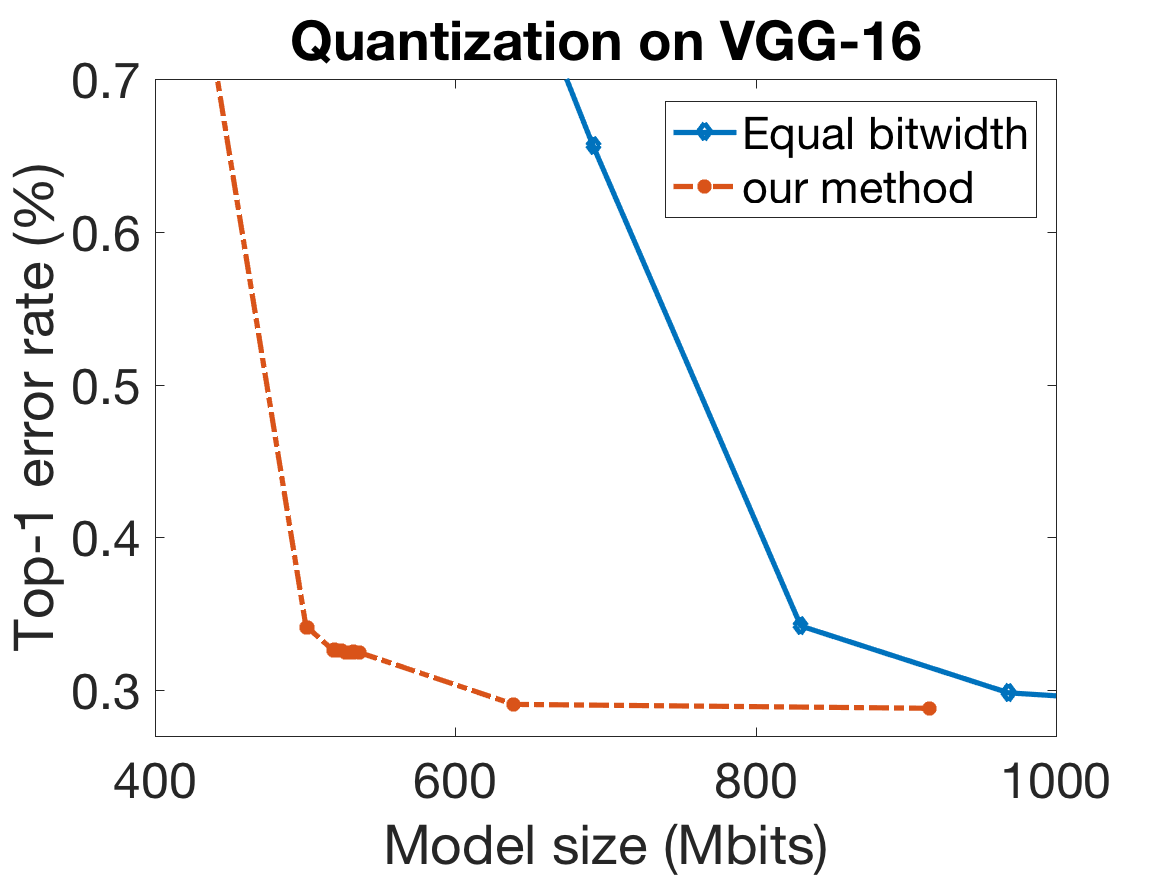}
    \label{fig:res_top1_compare_all_vgg}
}
\subfigure[]{
    \includegraphics[width=\figurewidthtwo\columnwidth]{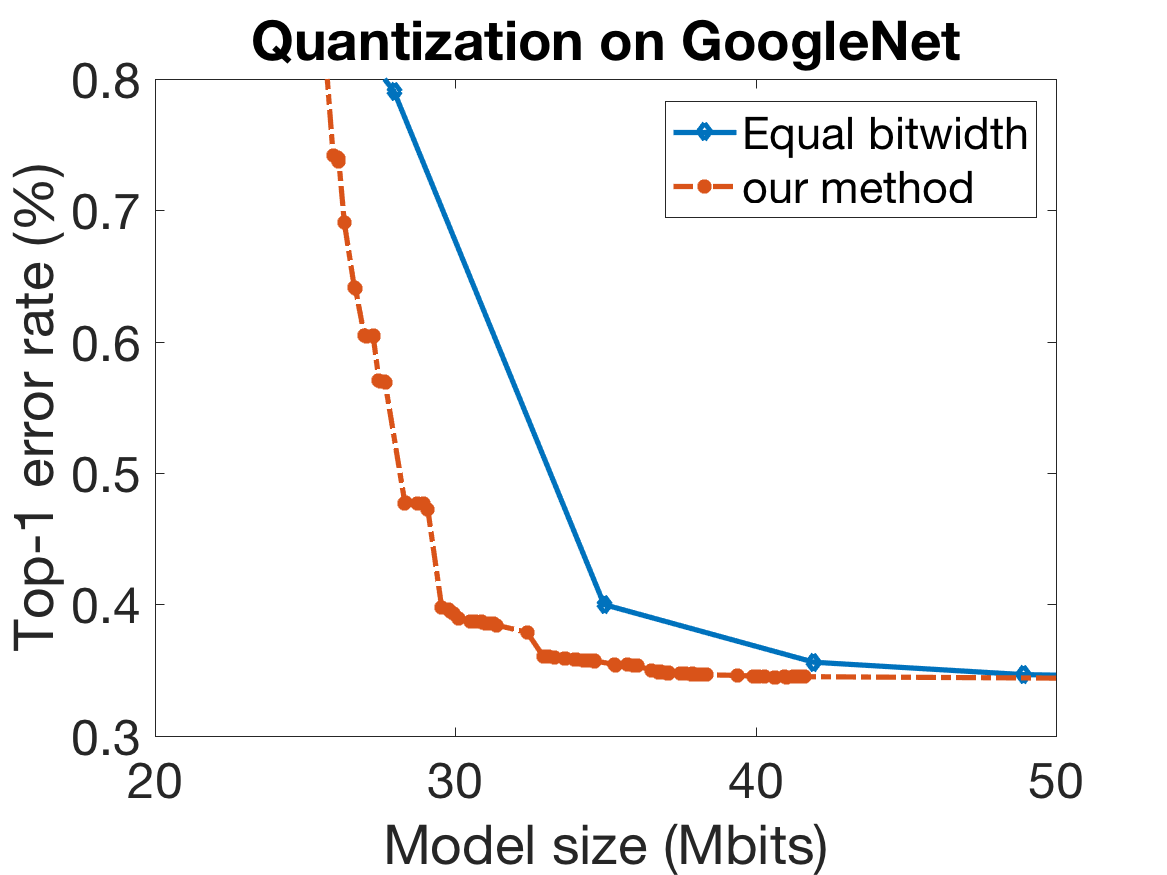}
    \label{fig:res_top1_compare_all_google}
}
\subfigure[]{
    \includegraphics[width=\figurewidthtwo\columnwidth]{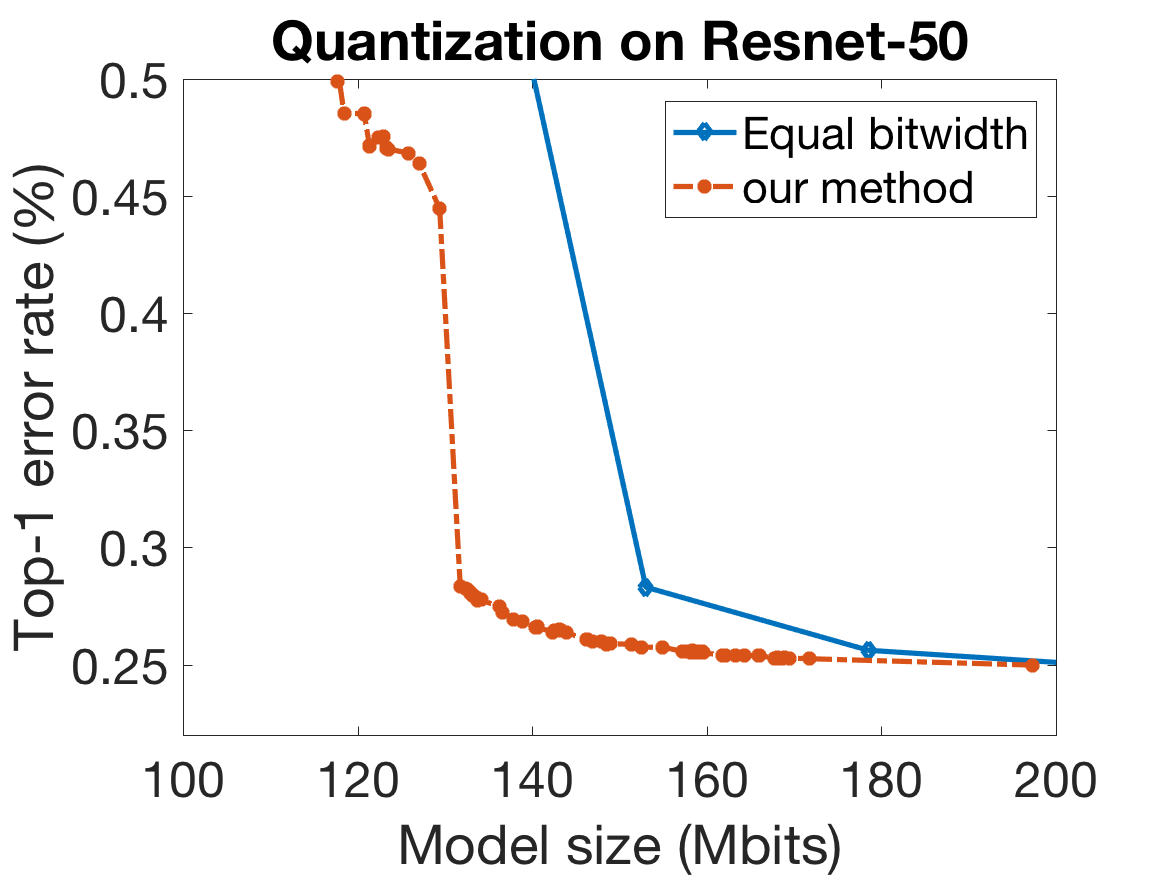}
    \label{fig:res_top1_compare_all_resnet}
}
\caption[]{Model size after quantization, v.s. model accuracy. All layers are quantized.}
\label{fig:res_top1_compare_all}
\end{figure}

Fig.~\ref{fig:res_top1_compare_all} shows the quantization results using our method, compared to equal quantization. For Alexnet and VGG-16 model, our method achieves $40\%$ less model size with same accuracy degradation, while for GoogleNet and Resnet-50, our method achieves $15-20\%$ less model size with same accuracy degradation. These results indicates that our proposed quantization method works better for models with more diverse layer-wise size and structures, like Alexnet and VGG.
